\documentclass[runningheads]{llncs}
\usepackage{float}       
\usepackage{graphicx}    
\usepackage{multirow}
\usepackage{pifont}
\usepackage[table]{xcolor}  
\usepackage[ruled,vlined]{algorithm2e}
\usepackage{algorithmic}
 
\usepackage[mobile]{eccv}

\usepackage{eccvabbrv}

\usepackage{graphicx}
\usepackage{booktabs}

\usepackage[accsupp]{axessibility}  

\usepackage{hyperref}

\usepackage{orcidlink}

\begin{document}

\title{Send Less, Perceive More: Masked Quantized Point Cloud Communication for Loss-Tolerant Collaborative Perception } 


\author{
Sheng Xu\inst{1} \and
Enshu Wang\inst{1}\inst{*} \and
Hongfei Xue\inst{2}\inst{*} \and
Jian Teng\inst{3} \and
Bingyi Liu\inst{3} \and
Yi Zhu\inst{4} \and
Pu Wang\inst{2} \and
Libing Wu\inst{1} \and
Chunming Qiao\inst{5}
}

\authorrunning{S.~Xu et al.}

\institute{
School of Cyber Science and Engineering, Wuhan University, China
\email{\{xusheng02, wanges17, wu\}@whu.edu.cn} \and
University of North Carolina at Charlotte, USA
\email{\{hongfei.xue, Pu.Wang\}@charlotte.edu} \and
Wuhan University of Technology, China
\email{\{tengjian, byliu\}@whut.edu.cn} \and
Wayne State University, USA
\email{yzhu39@wayne.edu} \and
University at Buffalo, USA
\email{qiao@buffalo.edu} \\
\inst{*}Corresponding authors.
}

\maketitle

\begin{abstract}
Collaborative perception enables connected vehicles to overcome occlusions and limited viewpoints by exchanging sensory information.
However, existing approaches face a fundamental trade-off between communication efficiency and perception accuracy, and remain highly vulnerable to random transmission packet loss.
We present QPoint2Comm, a quantized point-cloud communication framework that achieves high-fidelity 3D perception under strict bandwidth constraints while maintaining strong robustness to unreliable communication. 
Instead of transmitting high-dimensional intermediate features or raw point clouds, QPoint2-Comm encodes raw LiDAR measurements into compact discrete codebook indices and transmits only these indices. This discrete representation preserves explicit voxel-aligned geometric structure, avoiding secondary compression of already abstracted features and thus retaining richer spatial information at significantly lower bandwidth cost.
To further ensure robustness in real-world networks, we introduce a masked training strategy that explicitly simulates random packet loss during training.
By learning to reconstruct and reason from partially missing pillar features using a learnable feature filling mechanism, the model becomes inherently tolerant to severe transmission failures.
Extensive experiments on both simulated and real-world datasets demonstrate that QPoint2Comm sets a new state of the art in accuracy, communication efficiency, and resilience to packet loss.
\end{abstract}

\section{Introduction}
\label{sec:intro}

Multi-agent collaborative perception has emerged as a critical technology for autonomous driving systems~\cite{Li2022Bevformer,Levinson2011AutonomousDriving,Yang2023BevformerV2,Alotaibi2019LSAR,Chen2023CO3}. It enables multiple agents, such as vehicles and roadside units, to exchange sensor data (e.g., camera, LiDAR) in real time~\cite{Lu2023Robust3DDet,Redmon2016YOLO,Zhang2021SafeOcclusionAware}, extending each agent’s perceptual field beyond line-of-sight limitations.
By sharing information, collaborative perception effectively mitigates occlusion, sparse observations, and restricted viewpoints, resulting in improved accuracy and robustness for 3D object detection~\cite{Hu2022Where2comm,Chen20213DPointCloud,Hu2022Aerial3DDet}. With the rapid advancement of high-precision sensors and low-latency vehicular communication~\cite{Wang2020V2VNet,Chen20213DPointCloud}, collaborative perception has become a key enabler for long-range perception in connected and collaborative autonomous driving environments~\cite{Meng2023Hydro3D}.
\begin{figure}
  \centering
  \includegraphics[width=1.0\linewidth]{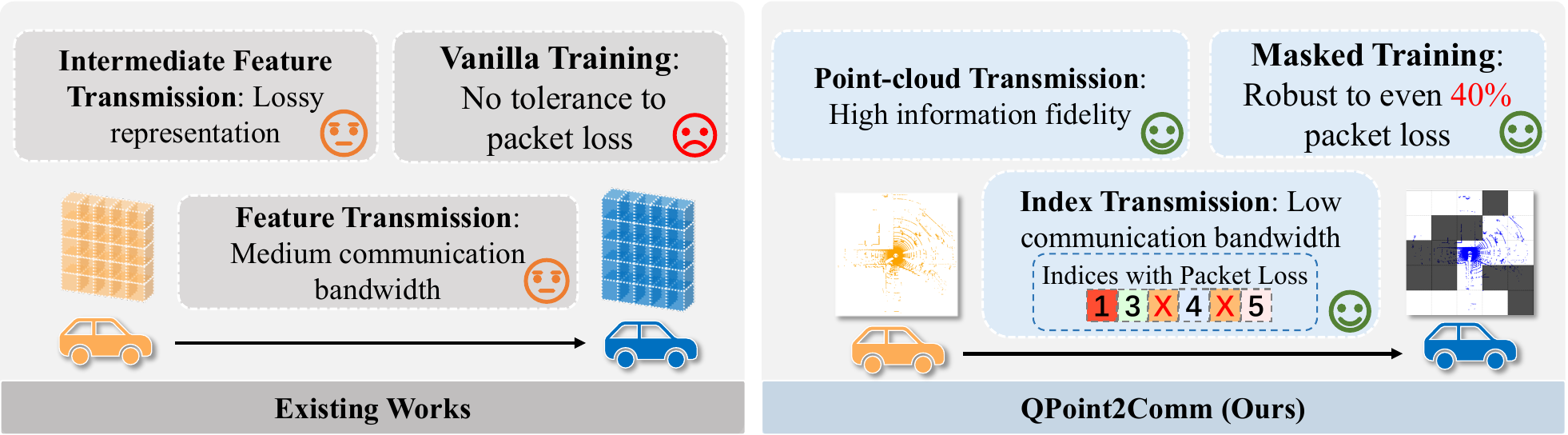}
  \vspace{-15pt}
  \caption{Comparison between our proposed quantized point cloud-based framework and existing feature-based methods for collaborative perception.}
  \label{fig:teaser}
  \vspace{-0.9cm} 
\end{figure}

Despite recent progress, practical deployment of collaborative perception still faces major challenges, including limited bandwidth~\cite{Wang2020V2VNet}, packet loss~\cite{Bian2024DistributedModel}, localization error~\cite{Lu2023Robust3DDet}, and transmission delay~\cite{Chen2023Transiff,Hu2022Where2comm}.
To \textbf{balance the bandwidth constraints and the perception performance}, existing methods can be categorized by the format of shared information:
(1) Raw-data-level transmission~\cite{Chen2019Cooper,Zhang2021EMP}, which achieves high accuracy but demands excessive bandwidth;
(2) Feature-level transmission~\cite{Liu2020When2com,Hu2022Where2comm,Zhang2024ERMVP,Xu2022V2XViT,Li2021DistilledGraph,Xu2022OPV2V}, which reduces bandwidth by sharing encoded features but still suffers from high communication volume and information loss;
(3) Result-level transmission~\cite{Shi2022VIPS,Song2023CoopPerception}, which shares only detection results (e.g., bounding boxes), offering high efficiency at the cost of losing contextual cues and increasing noise sensitivity.
While these approaches primarily focus on improving communication efficiency, real-world vehicular networks introduce an additional challenge: \textbf{communication instability}.
In practice, packet loss occurs unpredictably due to channel interference, dynamic topology, and environmental factors.
Even highly compressed representations can become unreliable once transmission is incomplete.
To mitigate packet loss, recent works either dynamically allocate communication resources~\cite{Chen2024Reinforcement,Yang2025UtilityAware,Fang2025RACP,Lin2024EdgeAssisted} or utilize historical frames to infer missing data~\cite{He2024RobustCollaborative,Tang2025RoCooper,Shi2025V2VCooperative}.
However, the former struggle to handle random loss patterns typical in real-world networks, while the latter accumulates prediction errors and introduces additional computational cost and latency.
Consequently, current frameworks lack inherent robustness to random packet loss during collaborative perception.
These limitations underscore two fundamental challenges in collaborative perception:
\textit{
(1) how to design a communication-efficient representation that preserves high-fidelity geometric information for accurate perception; and
(2) How to design a lightweight collaborative system that remains robust under unpredictable packet loss?
}

To address these challenges, we introduce QPoint2Comm, a collaborative perception framework that jointly addresses the efficiency–fidelity trade-off and enhances robustness to unreliable communication.
Our QPoint2Comm is built upon two key design ideas:
(1) Communicating compact discrete representations of raw point clouds instead of intermediate features, thereby preserving explicit geometric structure under strict bandwidth constraints; and
(2) Learning to perform collaborative perception under randomly masked transmissions, thereby achieving inherent tolerance to packet loss.
To realize the first idea, we introduce a Discrete Point Cloud Representation (DPR) that directly quantizes raw LiDAR measurements into compact discrete codebook indices. Unlike feature-level transmission, which already compresses and abstracts spatial information, QPoint2Comm communicates voxel-aligned discrete indices that preserve explicit geometric structure. This substantially reduces communication overhead while maintaining rich spatial information for accurate 3D perception.
To realize the second idea, we develop a masked training strategy that explicitly simulates random packet loss during training. By learning to reason from partially missing voxel features through a lightweight learnable feature filling mechanism, the model becomes inherently robust to incomplete transmissions, maintaining stable detection performance even under severe packet loss.
Beyond efficient and loss-tolerant communication, we further introduce a pyramid-scale cascade attention fusion (PCAF) module that reinforces ego features with filtered collaborative cues before multi-scale fusion. This design improves robustness to temporal delays and localization errors while enhancing detection accuracy. In addition, confidence-based feature filtering and bounding box refinement~\cite{Hu2022Where2comm,Liu2025mmCooper} further stabilize collaborative integration.
The main contributions of this work are summarized as follows:
\begin{itemize}
\item We propose \textbf{QPoint2Comm}, the first collaborative perception framework that transmits \emph{quantized point clouds} in a fully discrete form. By communicating compact voxel-aligned codebook indices instead of intermediate features, it significantly reduces bandwidth while preserving high-fidelity geometric structure for accurate 3D perception.
\item We introduce a \textbf{packet-loss-tolerant training paradigm} based on random masking and learnable feature filling. This design enables inherent robustness to unpredictable packet loss, allowing stable detection performance even under severe communication failures.
\item We develop a \textbf{pyramid-scale cascade attention fusion (PCAF)} strategy that reinforces ego features with filtered collaborative cues prior to multi-scale fusion, improving robustness to temporal delays, localization errors, and noisy transmissions.
\item Extensive experiments on OPV2V~\cite{Xu2022OPV2V} and DAIR-V2X~\cite{Yu2022DAIRV2X} demonstrate that QPoint2Comm consistently outperforms state-of-the-art methods, achieving AP@0.7 gains of 4.10\% and 5.18\%, respectively, while maintaining low bandwidth overhead and stable performance even under high packet loss rates.
\end{itemize}


\section{Related Work}
\label{sec:related}
\subsection{Collaborative Perception}
Collaborative perception enhances individual agents by sharing perception information to improve detection in complex environments. Existing methods are typically categorized by the type of transmitted data.
Raw-data-level transmission directly shares LiDAR point clouds, achieving strong performance but incurring prohibitive bandwidth costs~\cite{Chen2019Cooper,Zhang2023MultiVehicleCollab,Zhang2021EMP}. Although methods such as EMP~\cite{Zhang2021EMP} adaptively adjust the shared range, transmitting raw point clouds remains impractical under strict bandwidth constraints.
Conversely, result-level transmission shares only detection outputs, minimizing communication overhead but being sensitive to noise and localization errors~\cite{Xu2023MAMPerception,Shi2022VIPS,Huang2023V2XCoopPerception,Su20233DMultiObjectTracking}. Hybrid strategies like mmCooper~\cite{Liu2025mmCooper} combine multi-stage transmission, yet suffer from unstable efficiency. 
Differen from them, feature-level transmission shares encoded spatial features to reduce bandwidth~\cite{Singh2018LearnCommunicate,Liu2020When2com,Li2021DistilledGraph,Zhang2024ERMVP,Hu2022Where2comm,Xu2022V2XViT}. However, communication overhead is still substantial, and performance may degrade due to information loss during encoding. To alleviate this, recent works such as Where2comm~\cite{Hu2022Where2comm} selectively transmit high-confidence features.
In order to further reduce bandwidth, CodeFilling~\cite{YueCodeFilling:CVPR2024}, CoGMP~\cite{Fu_2025_CVPR}, and QCTF~\cite{Chen2025QCTF} apply VQ-VAE~\cite{oord2017vqvae}-based quantization on already extracted features. Since feature extraction itself compresses the raw point cloud, such feature-level quantization effectively performs a second-stage compression, inevitably introducing additional information loss.
\textit{In contrast, our method fundamentally differs from existing feature-level approaches, including VQ-VAE~\cite{oord2017vqvae}-based quantization methods.} First, we quantize raw LiDAR rather than latent semantic features, avoiding secondary compression. Second, we reconstruct geometrically structured point clouds instead of feature maps without explicit spatial form. Third, our transmitted discrete indices are voxel-aligned and spatially grounded, carrying explicit geometric meaning, whereas feature-level methods transmit feature-semantic codes without direct geometric correspondence.
Moreover, through masked training, the framework achieves inherent tolerance to packet loss, while cascading ego features prior to fusion further improves robustness and detection performance.

\subsection{Point Cloud Representation}
LiDAR point clouds~\cite{Girdhar2016VectorRepresentation} capture 3D geometry and distance. Early methods like LiDAR GAN~\cite{Caccia2019LidarGen}, LiDAR VAE~\cite{Caccia2019LidarGen}, and ProjectedGAN~\cite{Sauer2021ProjectedGANs} synthesize point clouds from partial inputs but struggle to reconstruct attributes like intensity. Diffusion-based approaches (R2DM~\cite{nakashima2024lidar}, LiDARGen~\cite{Zyrianov2022LidarGen}, LiDM~\cite{ran2024towards}) generate high-fidelity clouds but focus on generation and require iterative steps, limiting real-time use. VQ-VAE~\cite{oord2017vqvae}-based methods such as Point-BERT~\cite{yu2021pointbert} and UltraLiDAR~\cite{xiong2023learning} enable compact encoding and reconstruction but often struggle with sparsity or attribute recovery. Motivated by this, we propose a discrete LiDAR representation that supports compact encoding, accurate one-step reconstruction, and low-bandwidth transmission while preserving geometry and intensity for collaborative perception.

\subsection{Packet Loss during Transmission}
To address communication unreliability in collaborative perception, several recent studies have focused on mitigating packet loss. Such as RL-based~\cite{Chen2024Reinforcement} and UAHRL~\cite{Yang2025UtilityAware}, mitigate packet loss by dynamically allocating communication resources. R-ACP~\cite{Fang2025RACP} and Edge-Assisted~\cite{Lin2024EdgeAssisted} address the issue by selectively discarding less important information. PACP~\cite{Fang2024PACP} alleviates packet loss by discarding redundant features and dynamically compressing critical ones.
SmartCooper~\cite{Zhang2024SmartCooper} enhances robustness by adaptively dropping negatively contributing data. However, the above methods struggle to handle the random loss that occurs in real-world scenarios.
To overcome this limitation, another line of research explores approaches that can handle random packet loss. Methods such as RCooper~\cite{Tang2025RoCooper} and RCP~\cite{He2024RobustCollaborative} leverage historical frames, but errors accumulate when multiple frames are lost. V2V-Cooper~\cite{Shi2025V2VCooperative} and LCRN~\cite{Li2023V2VPerception} use ego information to recover missing regions or enhance detection performance but incur high computational cost.
These methods degrade significantly under high packet loss and lack inherent loss tolerance, limiting practical deployment. In contrast, we adopt a lightweight random masking strategy that enables inherent tolerance to real-world packet loss without relying on historical frames, achieving robust perception under severe loss with low computational overhead.

\begin{figure*}[tbh]
\vspace{-0.5cm}
  \centering
  \includegraphics[width=1.0\linewidth]{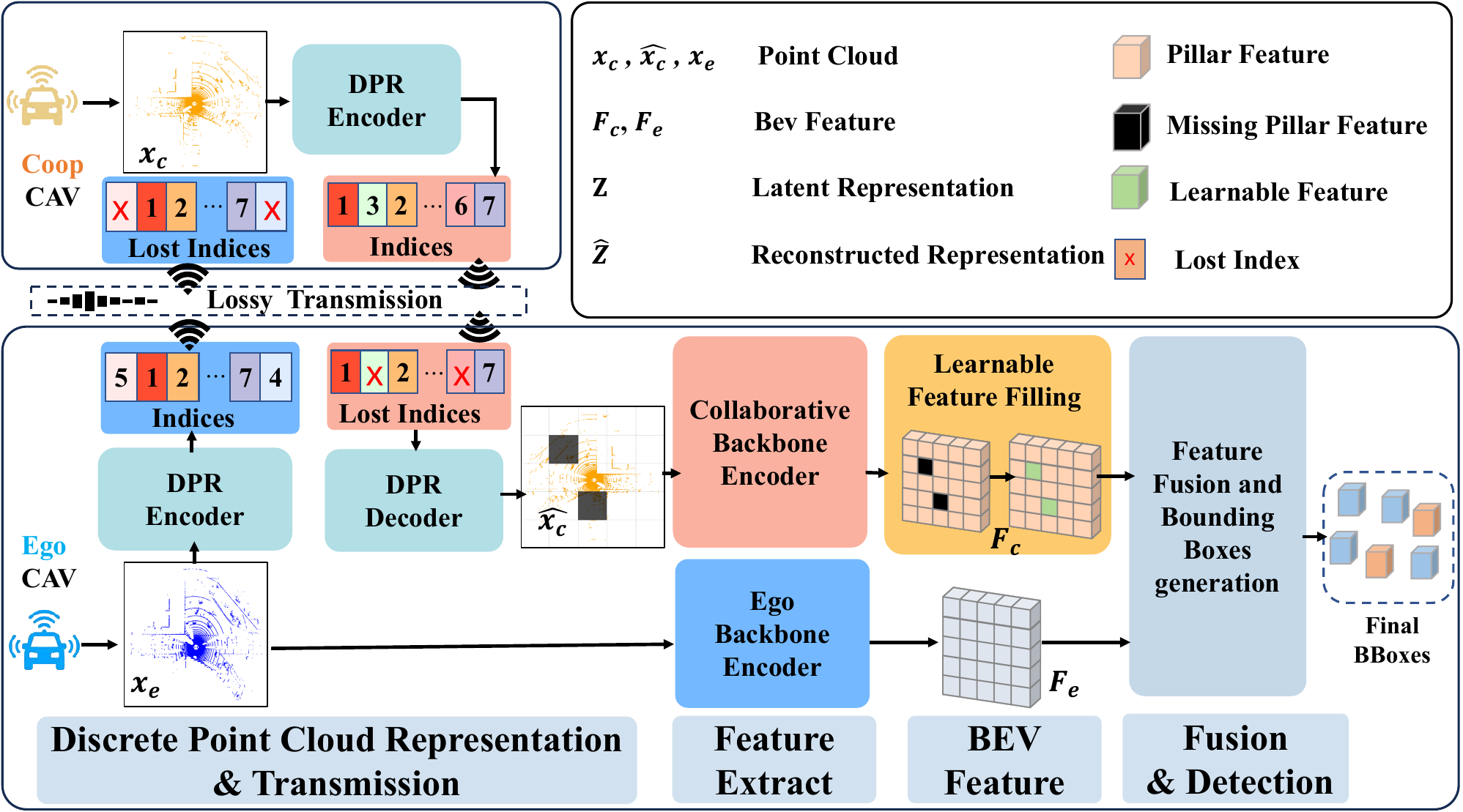}
  \caption{Overview of the proposed framework, consisting of Discrete Point Cloud Representation (\cref{subsec:DPR}) for point cloud encoding and decoding, Packet-loss Tolerant Design (\cref{subsec:PTD}) to enable communication with missing packets, Feature Fusion and Bounding Boxes Generation (\cref{subsec:FFBBG}) module for generating bounding boxes.}
  \label{fig:framework}
  \vspace{-1.2cm} 
\end{figure*}

\section{Methodology}
\label{sec:methodology}
\subsection{Overview}
\label{subsec:overview}
Conventional collaborative perception methods based on feature-level sharing struggle to preserve explicit geometric structure under strict bandwidth constraints and are vulnerable to random packet loss. As illustrated in \cref{fig:teaser}, our method insteadly communicates compact discrete representations of raw point clouds rather than intermediate features, retaining explicit spatial geometry at low bandwidth. Furthermore, by training with randomly masked strategy, the model learns to perform collaborative perception under incomplete communication, achieving inherent tolerance to packet loss while maintaining accurate detection.

To send less while preserving more information and remaining robust to packet loss, we propose QPoint2Comm, a collaborative perception framework that encodes LiDAR point clouds into compact discrete indices via a shared codebook for communication and reconstructs point representations on the ego vehicle.
As illustrated in \cref{fig:framework}, each collaborative agent first uses the DPR Encoder (\cref{subsec:DPR}) to convert raw LiDAR data into semantically discrete codes, significantly reducing bandwidth. The ego vehicle then applies the DPR Decoder to reconstruct collaborative point clouds, retaining detailed spatial structure.
To address packet loss, missing features are replaced with a learnable representation (\cref{subsec:PTD}), enabling inherent tolerance to incomplete transmissions. Finally, the Feature Fusion and Bounding Box Generation module (\cref{subsec:FFBBG}) generates detection results, where the Pyramid-scale Cascade Attention Fusion (PCAF) module (\cref{subsec:PCAF}) produces the fused feature.

\subsection{Discrete Point Cloud Representation}
\label{subsec:DPR}

\begin{figure}[H]
\vspace{-0.65cm}
  \centering
  \includegraphics[width=0.85\linewidth]{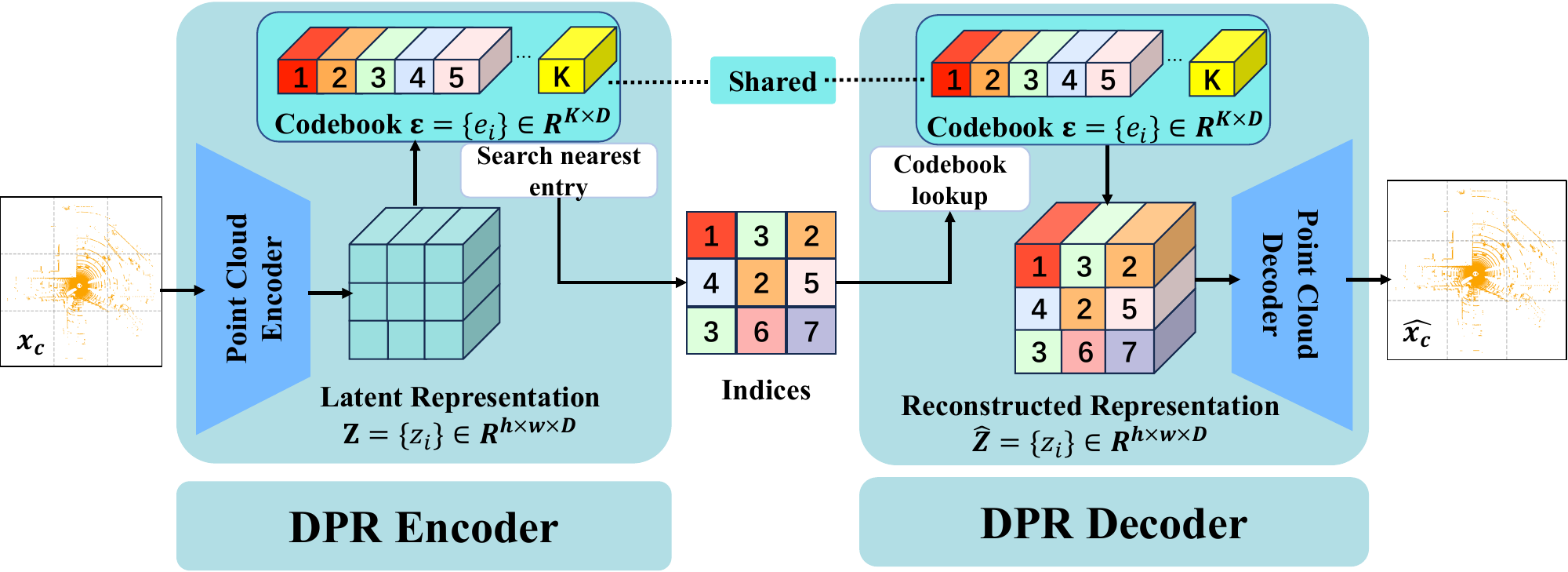}
  \vspace{-5pt}
  \caption{The Discrete Point Cloud Representation (DPR) module is trained to encode point clouds into discrete index sequences and to faithfully reconstruct them via a shared codebook.}
  \label{fig:DPR}
  \vspace{-20pt} 
\end{figure}

\noindent
As shown in \cref{fig:DPR}, the Discrete Point Cloud Representation (DPR) module encodes LiDAR point clouds into compact, transmission-efficient discrete codes using shared codebooks, ensuring consistent representation across collaborative and ego agents. Unlike standard VQ-VAE~\cite{oord2017vqvae}, which struggles with sparse and continuous 4D point cloud distributions (3D coordinates and reflectivity), DPR adopts a one-hot voxel grid strategy. The input point cloud is discretized into a regular 3D voxel grid of size $\mathcal{H} \times \mathcal{W} \times \mathcal{L}$, analogous to 2D pixel grids. To represent both geometry and reflectivity, DPR employs a dual-codebook design:
\begin{itemize} 
\item a spatial occupancy codebook to model grid-level occupancy patterns;
\item an intensity codebook to capture grid-wise reflectivity characteristics. 
\end{itemize}
The DPR module is composed of three core components:
\textbf{(1) Point Cloud Encoder $E$}:
Given a point cloud $x$, we first voxelize it into a 3D grid and define an occupancy tensor 
${X}_{\text{occ}} \in \{0,1\}^{\mathcal{H} \times \mathcal{W} \times \mathcal{L}}$ 
and an intensity tensor 
${X}_{\text{int}} \in [0,1]^{\mathcal{H} \times \mathcal{W} \times \mathcal{L}}$.
These are passed through corresponding encoders to obtain latent representations:
${Z} = E({X}) \in \mathbb{R}^{h \times w \times D}$.
\textbf{(2) Vector Quantization}:
Each vector ${z}_{i}$ in ${Z}$ is replaced with the nearest entry in the learnable codebook $\mathcal{E}=\{{e}_k \}_{k=1}^{K}$, where ${e}_k \in \mathbb{R}^D$. The quantized resonstructed representation is denoted as: $\hat{{Z}}_{\text{}} = q_{\text{}}({Z}_{\text{}}) = \{{e}_{k_{\text{}}^*}^{\text{}}\} $, where $k_{}^*$ is computed by: $k_{}^* = \arg\min_k \|{z}_{i}^{\text{}} - {e}_k^{\text{}}\|_2^2$. 
\textbf{(3) Point Cloud Decoder}:
To reconstruct the point cloud, we sample points from Gaussian distributions centered at the grid centroids. The decoder is denoted as:
$\hat{X} = \text{Dec}(\hat{Z}; \mathcal{N}(\mu, \sigma^2 I)),$
where \(\mu\) denotes the grid centroid, and \(\sigma^2\) controls the variance of the Gaussian sampling around the centroid. This stochastic reconstruction enhances robustness by capturing the point distribution within each occupied grid.

\noindent \textbf{Training.} The DPR module is trained using the following losses. The quantization losses for the occupancy codebook $\mathcal{L}_{\text{vq,occ}}$ and the intensity codebook $\mathcal{L}_{\text{vq,int}}$ are both computed as follows:
$\mathcal{L}_{\text{vq}} = \| {X} - \hat{{X}} \|_2^2 + \| \text{sg}[E({X})] - \hat{{Z}} \|_2^2 + \| \text{sg}[\hat{{Z}}] - E({X}) \|_2^2,$
where $\text{sg}[\cdot]$ denotes the stop-gradient operation~\cite{oord2017vqvae}. Occupancy Reconstruction Loss is computed using voxel-wise binary cross-entropy:
$\mathcal{L}_{\text{occ\_reconstruct}} = - \frac{1}{V} \sum_{v=1}^{V} \Big[ y_v \log(\hat{y}_v) + (1 - y_v) \log(1 - \hat{y}_v) \Big],$
where $y_v \in \{0,1\}$ denotes the ground-truth occupancy label, $V$ is the total grids and $\hat{y}_v$ is the predicted probability. Reflectivity Intensity Loss is computed as mean squared error over all occupied grids:
$\mathcal{L}_{\text{int\_reconstruct}} = \frac{1}{N_{\text{occ}}} \sum_{v=1}^{N_{\text{occ}}} (\hat{I}_v - I_v^{\text{gt}})^2,$
where $\hat{I}_v$ and $I_v^{\text{gt}}$ are the predicted and ground-truth normalized intensities, the $N_{\text{occ}}$ denotes the number of truly occupied grids.

As shown in \cref{fig:framework}, the DPR module enables collaborative agents to transmit compact codebook indices to the ego agent. Ego reconstructs the point cloud via the shared codebooks, allowing low-bandwidth communication while preserving rich information, improving detection accuracy.

\subsection{Packet-loss Tolerant Design}
\label{subsec:PTD}
\textbf{Detection Backbone.} In multi-agent collaborative perception, let ${N} = \{1, \ldots, n\}$ denote the set of agents. When agent $i$ is the ego vehicle, the remaining $n-1$ agents transmit their discrete indices and poses. The ego agent reconstructs and transforms the received point clouds into its coordinate frame. Our model uses distinct PointPillar~\cite{Lang2019PointPillars} backbones for the ego and collaborators, denoted as $\psi_E$ and $\psi_C$. The ego feature is encoded into a BEV representation $F_e = \psi_E(x_e) \in \mathbb{R}^{C \times H \times W}$, while each collaborator’s feature is $F_c = \psi_C(x_c)$, where $x_e$ and $x_c$ are the point clouds, and $C$, $H$, $W$ are the feature map dimensions.

\noindent \textbf{Mask Training.} 
At the ego agent, the point clouds received from collaborative agents are processed by the collaborative
backbone encoder $\psi_C(\cdot)$ to extract pillar features. To improve robustness against packet loss, a random masking function $\mathcal{M}_r(\cdot)$ is applied on the extracted pillar features during training:
$\tilde{F} = \mathcal{M}_r\big(\psi_C(x_{\text{c}})\big) \in \mathbb{R}^{C \times H \times W},$
where $\tilde{F}$ contains randomly masked features. For masked regions, their features are replaced by a learnable feature  $f_{\text{learnable}} \in \mathbb{R}^{C}$, yielding the collaborative feature $F_c$.
This Mask Training design allows the model to tolerate random packet loss in real-world communication, as the learnable feature effectively fills missing pillar features, making the collaborative feature robust to incomplete transmissions. 

As shown in \cref{fig:framework}, during inference, some codebook indices may be lost due to environmental fluctuations, leading to missing pillar features. These missing features are replaced with the learnable feature $f_{\text{learnable}}$, producing the collaborative feature $F_c$.
In practice, one codebook index corresponds to multiple voxel grids, the loss of a single codebook index can result in partial occupancy loss across multiple voxel grids. We adopt a strategy where any grid with missing portions is treated as fully lost. 
Mask Training randomly masks features and fills them with a learnable feature $f_{\text{learnable}}$, enabling inherent tolerance to packet loss and robust perception under high-loss conditions.

\subsection{Feature Fusion and Bounding Boxes Generation}
\label{subsec:FFBBG}
\begin{figure}[h]
\vspace{-1.45cm}
  \centering
  \includegraphics[width=0.9\linewidth]{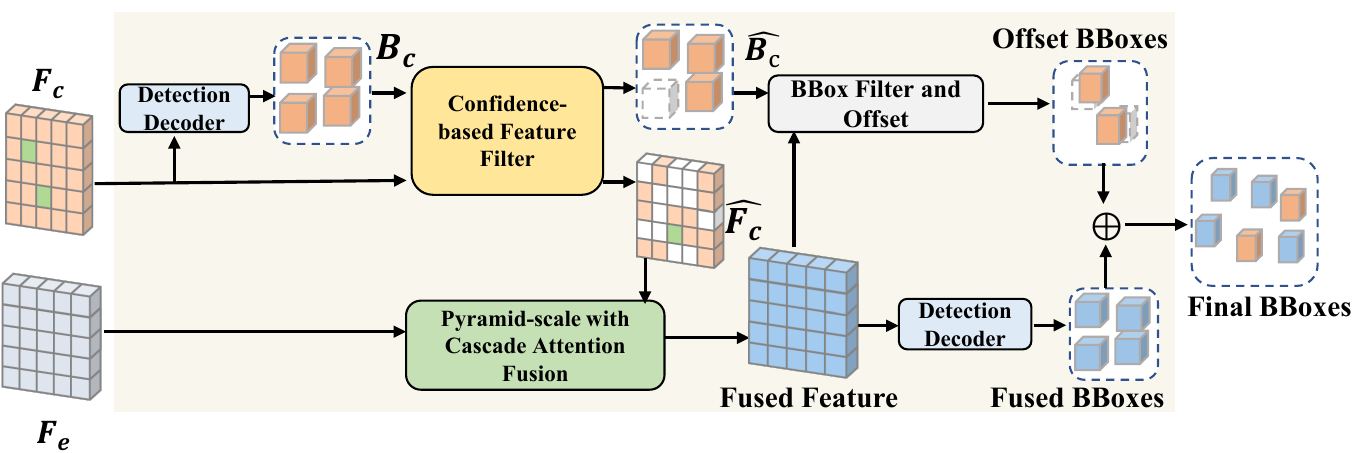}
  \vspace{-15pt}
  \caption{The Feature Fusion and Bounding Boxes generation. }
  \label{fig:FFBBG}
  \vspace{-15pt} 
\end{figure}

\noindent As illustrated in \cref{fig:FFBBG}, both collaborative feature $F_c$ and ego feature $F_e$ are fed into the Feature Fusion and Bounding Box Generation module, which produces the final detection results. The key components of this module are as follows:\\
\textbf{Confidence-based Feature Filter.} 
Motivated by the design in Where2comm~\cite{Hu2022Where2comm}, the Confidence-based Feature Filtering (CFF) module suppresses low-confidence regions. Each $F_c$ is processed by a network to produce spatial confidence maps ${G}_f, {G}_b \in [0,1]^{H \times W}$. Regions with confidence below the $p$-th percentile $\tau_p$ are filtered, and Gaussian smoothing~\cite{Maddison2016Concrete} reduces noise. The filtered confidence maps are computed as:
$\mathbb{I}_{\{G_{f,b} > \tau_p\}} = 1 \text{ if } G_{f,b} > \tau_p, \text{ otherwise } 0,$
\label{eq:indicator_function}
$\hat{G}_{f,b} = G_{f,b} \odot \mathbb{I}_{\{G_{f,b} > \tau_p\}} \odot \mathcal{G}_{\sigma}(G_{f,b})$, and the filtered collaborative features and bounding boxes are obtained by $\hat{F}_c = \hat{G}_f \otimes F_c$ and $\hat{B}_c = \hat{G}_b \otimes B_c$, where $\mathbb{I}$ denotes the indicator function, ${{B}}_c = \text{Detector}(F_c)$ and $\mathcal{G}_{\sigma}[\cdot]$ is a Gaussian smoothing filter.\\
\textbf{Pyramid-scale Fusion with Cascade Attention.}
The Pyramid-scale Fusion with Cascade Attention module generates fused features $\mathcal{F}_i$ used both to produce the fused bounding boxes and to refine collaborative bounding boxes within the BFO module, which will be elaborated in~\cref{subsec:PCAF}\\
\textbf{BBox Filter and Offset.}
Following the approach of mmCooper~\cite{Liu2025mmCooper}, the BBox Filter and Offset (BFO) refines collaborative bounding boxes $\hat{B}_c$. Low-quality boxes are discarded, and remaining boxes are corrected. Offsets are supervised by smooth absolute error $\mathcal{L}_{{off}}$~\cite{Girshick2015FastRCNN}, and quality scores by focal loss $\mathcal{L}_{score}$~\cite{Lin2017FocalLoss}.

\subsection{Pyramid-scale Fusion with Cascade Attention}
\label{subsec:PCAF}
\begin{figure}
\vspace{-5pt}
  \centering
  \includegraphics[width=0.8\linewidth]{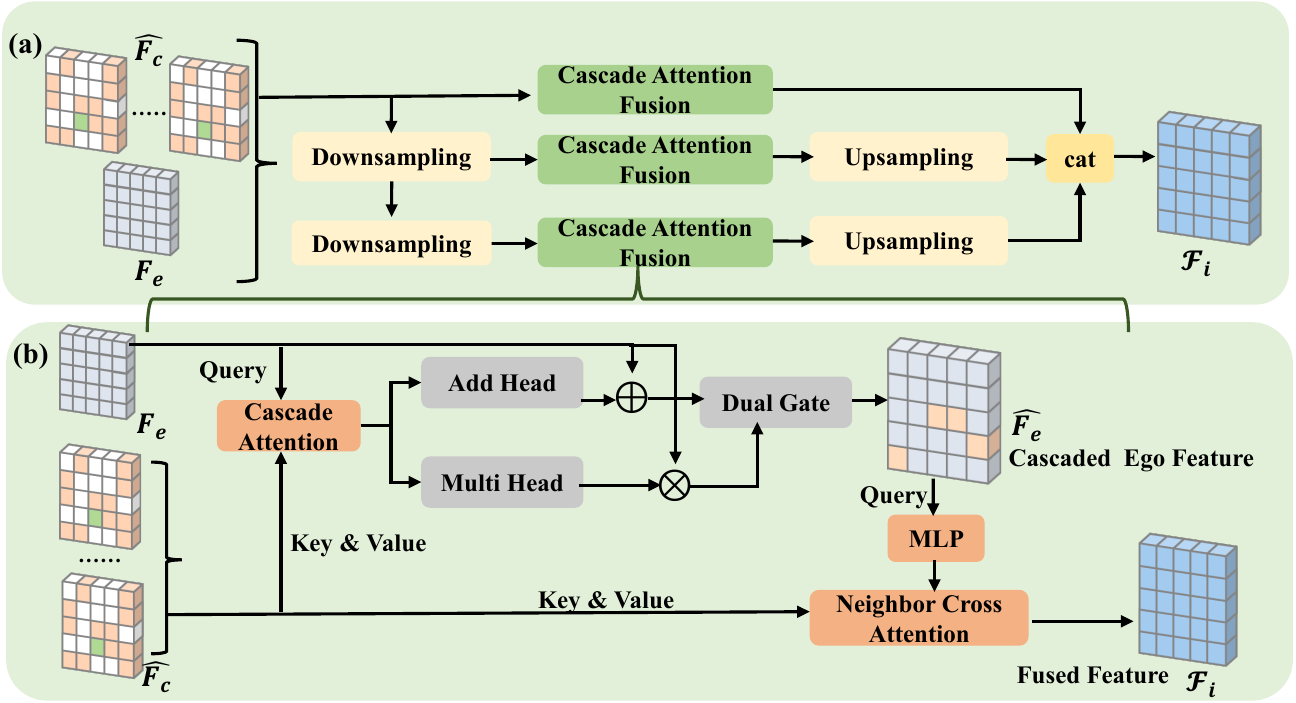}
  \vspace{-5pt}
  \caption{(a) The Pyramid-scale Fusion. (b) The Cascade Attention Fusion (CAF).}
  \vspace{-20pt}
  \label{fig:PCAF}
\end{figure}

\noindent As illustrated in \cref{fig:PCAF}, the Pyramid-scale Cascade Attention Fusion (PCAF) module employs Pyramid-scale Feature Fusion (PFF) to aggregate multi-resolution context. At each scale, Cascade Attention Fusion (CAF) sequentially fuses ego features with filtered collaborative features, enabling complementary information integration, alleviating occlusion and uncertainty, and producing robust representations for detection.\\
\textbf{Cascade Attention Fusion (CAF).} Within the CAF, the ego features serve as Query while the collaborative features provide Key and Value. The CAF output is split into two augmentation pathways to supplementary and enhancement effects simultaneously:\\
(1) Additive path for information supplement, which introduces complementary cues from collaborators:
$A_{\text{add}} = H_{\text{add}}(\text{Attn}(F_e, \hat{F}_c, \hat{F}_c)) + F_e,$\\
(2) Multiplicative path for information enhancement, which enhances ego representation based on the collaborative cues:
$A_{\text{mul}} = H_{\text{mul}}(\text{Attn}(F_e, \hat{F}_c, \hat{F}_c)) \otimes F_e,$\\
These two paths are then dynamically fused via a learnable gating mechanism to balance information supplement and information enhancement:
$\hat{F}_e = \sigma(W_g) \otimes A_{\text{add}} + (1 - \sigma(W_g)) \otimes A_{\text{mul}}$.

\noindent where $H_{\text{add}}$ and $H_{\text{mul}}$ denote projection heads for each path, $\sigma$ is the sigmoid function, and $W_g$ are channel-wise gating weights that adaptively balance their contributions. This design reinforces ego features before pyramid-scale fusion, improving robustness and precision.
The PFF then fuses cascaded ego features $\hat{F}_e$ and filtered collaborative features $\hat{F}_c$ across pyramid levels via neighbor cross-attention, followed by upsampling and concatenation to produce the final fused feature $\mathcal{F}_i$.
The CAF module cascades ego features before fusing with pyramid-scale collaborative information, improving robustness and accuracy compared to direct fusion.


\subsection{Loss Functions}
\textbf{Training Stage 1: Discrete Point Cloud Representation.} The occupancy and intensity codebooks are trained separately with losses 
\(\mathcal{L}_{\text{occ}} = \mathcal{L}_{\text{vq,occ}} + \mathcal{L}_{\text{occ\_reconstruct}}\), 
\(\mathcal{L}_{\text{int}} = \mathcal{L}_{\text{vq,int}} + \mathcal{L}_{\text{int\_reconstruct}}\), 
where \(\mathcal{L}_{\text{vq,occ}}\) and \(\mathcal{L}_{\text{vq,int}}\) denote the VQ losses for occupancy and intensity codebooks.\\
\textbf{Training Stage 2: Object Detection.} The detection heads~\cite{Carion2020DETR} decode the fused features $\mathcal{F}_i$. The regression head predicts geometric parameters for each anchor box:
$\mathcal{O}_{\text{reg}} = f_{\text{dec}}^r(\mathcal{F}_i) \in \mathbb{R}^{7 \times H \times W}$, where the output channels correspond to the center coordinates $(x, y, z)$, box dimensions $(l, w, h)$, and yaw angle $\theta$. The classification head produces confidence scores:
$\mathcal{O}_{\text{cls}} = f_{\text{dec}}^c(\mathcal{F}_i) \in \mathbb{R}^{2 \times H \times W}$, The overall detection loss is formulated as:
$\mathcal{L} = \mathcal{L}_{\text{reg}} + \mathcal{L}_{\text{cls}} + \mathcal{L}_{\text{off}} + \mathcal{L}_{\text{score}},
\label{eq:total_loss_1}
$
where $\mathcal{L}_{\text{reg}}$ denotes the regression loss for geometric parameters, $\mathcal{L}_{\text{cls}}$ is the focal loss for classification.

\section{Experiments}
\label{sub:experiments}

\subsection{Datasets and Experimental Settings}
\textbf{Datasets.} OPV2V~\cite{Xu2022OPV2V} is a public vehicle-to-vehicle collaborative perception dataset, simulated by OpenCDA~\cite{Xu2021OpenCDA} and CARLA~\cite{Dosovitskiy17}, containing 73 driving scenarios with synchronized LiDAR point clouds and RGB images, and over 11,464 annotated frames encompassing more than 230,000 precise 3D bounding boxes. DAIR-V2X~\cite{Yu2022DAIRV2X} is a large-scale real-world multimodal dataset for 3D object detection, comprising over 71,254 synchronized frames of LiDAR point clouds and camera images.\\
\textbf{Evaluation Metrics.} We report Average Precision at IoU thresholds~\cite{Everingham2010The} of 0.5 and 0.7, following the KITTI~\cite{Menze2015ObjectSceneFlow} benchmark. Communication cost is measured by:
$\log_2\big((2\times\mathcal{N} \times \log_2 K + 6 \times 32) / 8\big),$
where $\mathcal{N}=h \times w$ is the number of quantized vectors, $K$ is the codebook size, 2 represents dual-coodbook, and 6 represents 6 pose-related parameters (32 bits each), converted to bytes.\\
\textbf{Implementation Details.} 
All models are trained on NVIDIA L20 GPUs using a two-stage pipeline. Stage 1 pretrains DPR for 160 epochs (batch size 3, codebook size 2,048, vector dimension 1,024). Stage 2 trains a PointPillar~\cite{Lang2019PointPillars} backbone for 60 epochs, with batch size of 3 on OPV2V and 5 on DAIR-V2X. The mask ratio is 0.3 for both datasets, CFF percentiles $p$-th are set to 0.35 (OPV2V) and 0.2 (DAIR-V2X), and the codebook is frozen after Stage 1.
\subsection{Quantitative Results}

\begin{figure}[t]
  \centering
  \captionsetup{font=small, skip=2pt}
  \begin{minipage}[t]{0.5\linewidth}
    \centering
    \begin{table}[H]
      \centering
      \caption{Perception Performance on OPV2V and DAIR-V2X. Conditions: 100ms Time Delay, 0.2m Localization Error, 0.2° Heading Error. Bold = Best, Underline = Second-Best.}
      \label{tab:main results}
      \resizebox{1.0\linewidth}{!}{%
      \renewcommand{\arraystretch}{1.7}
      \begin{tabular}{c|c|c}
      \hline
      \multirow{2}{*}{\textbf{Models}} & \multicolumn{1}{c|}{\textbf{OPV2V}} & \multicolumn{1}{c}{\textbf{DAIR-V2X}} \\ 
      \cline{2-3}
       & \textbf{AP@0.7 / AP@0.5 } & \textbf{AP@0.7 / AP@0.5 } \\ 
      \hline
      No Fusion~\cite{Lang2019PointPillars}       & 48.66/68.71 & 43.57/50.03 \\ 
      \hline
      Late Fusion~\cite{Lang2019PointPillars}     & 59.48/79.62 & 34.47/51.14 \\ 
      \hline
      Intermediate Fusion~\cite{Lang2019PointPillars}  & 70.82/88.41 & 39.38/56.22 \\ 
      \hline
      When2comm~\cite{Liu2020When2com}          & 57.55/74.11 & 33.68/48.20 \\ 
      \hline
      DisoNet~\cite{Li2021DistilledGraph}           & 68.64/84.72 & 40.69/52.67 \\ 
      \hline
      Where2comm~\cite{Hu2022Where2comm}        & 69.73/85.16 & 43.71/59.52 \\ 
      \hline
      V2X-ViT~\cite{Xu2022V2XViT}         & 70.06/84.65 & 40.43/53.08 \\ 
      \hline
      ERMVP~\cite{Zhang2024ERMVP}            & 69.71/86.63 & 46.96/64.21 \\ 
      \hline
      SICP~\cite{Qu2024SICP}            & 67.13/82.86 & 41.03/52.72 \\ 
      \hline
      mmCooper~\cite{Liu2025mmCooper}          & \underline{78.11}/\underline{88.93} & \underline{48.27}/\underline{65.12} \\ 
      \hline
      \textbf{Ours}         & \textbf{82.21}/\textbf{92.18} & \textbf{53.45}/\textbf{67.97} \\ 
      \hline
      \end{tabular}
      }
      \vspace{-3mm} 
    \end{table}
  \end{minipage}
  \hfill 
  \begin{minipage}[t]{0.47\linewidth}
    \centering
    \begin{figure}[H]
      \centering
      \includegraphics[width=1.1\linewidth, keepaspectratio]{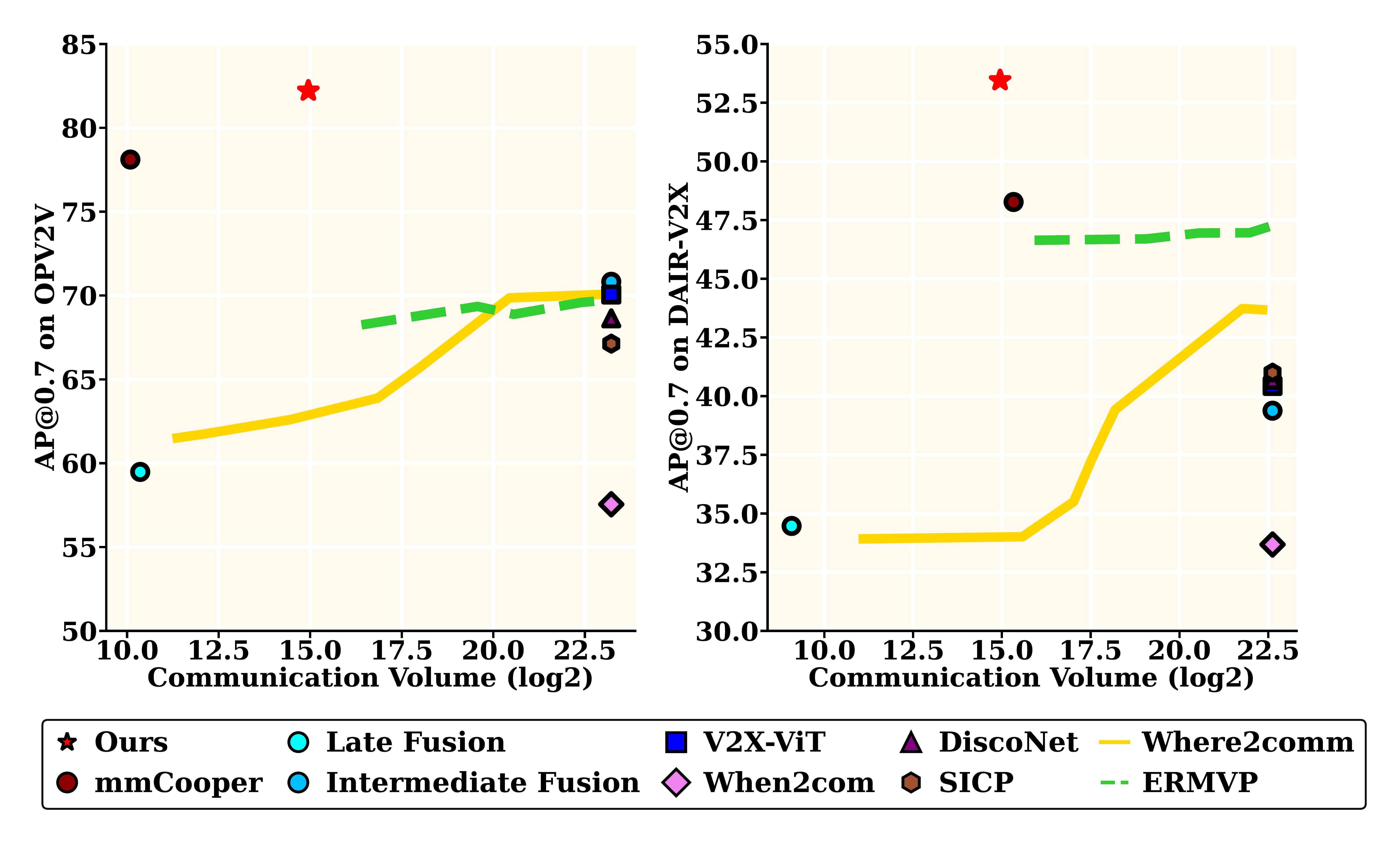}
      \renewcommand{\arraystretch}{1.8}
      \caption{Dection performance and communication volumes on the OPV2V and DAIR-V2X.}
      \label{fig:main result}
      \vspace{-4mm} 
    \end{figure}
    \vspace{-11mm} 

    \begin{table}[H]
      \centering
      \caption{Comparison with CodeFilling on OPV2V.}
      \vspace{-4mm} 
      \label{tab:comparison_codefilling_opv2v}
      \resizebox{1.1\linewidth}{!}{%
      \renewcommand{\arraystretch}{1.6}
      \begin{tabular}{c|c|c|c|c}
      \hline
      Method & \begin{tabular}{@{}c@{}}\textbf{0 ms}\\[2pt]\textbf{AP@0.5 / AP@0.7}\end{tabular}
             & \begin{tabular}{@{}c@{}}\textbf{100 ms}\\[2pt]\textbf{AP@0.5 / AP@0.7}\end{tabular}
             & \begin{tabular}{@{}c@{}}\textbf{200 ms}\\[2pt]\textbf{AP@0.5 / AP@0.7}\end{tabular}
             & \begin{tabular}{@{}c@{}}\textbf{300 ms}\\[2pt]\textbf{AP@0.5 / AP@0.7}\end{tabular} \\
      \hline
      Ours        & \textbf{92.55} / 86.81  & \textbf{90.14} / \textbf{81.78}  & \textbf{90.80} / \textbf{80.56}  & \textbf{89.73} / \textbf{80.13}  \\
      \hline
      CodeFilling & 90.82 / \textbf{88.19}  & 85.29 / 59.63  & 75.16 / 58.67  & 73.69 / 57.13 \\
      \hline
      \end{tabular}
      }
      \vspace{-3mm} 
    \end{table}
  \end{minipage}

  \vspace{-3mm}
\end{figure}

\textbf{Detection Precision.} As shown in \cref{tab:main results}, under  conditions with 100\,ms time delay, 0.2\,m localization error, and 0.2° heading error, our proposed QPoint2Comm consistently outperforms all baselines (When2com~\cite{Liu2020When2com}, DisoNet~\cite{Li2021DistilledGraph}, Where2comm~\cite{Hu2022Where2comm}, V2X-ViT~\cite{Xu2022V2XViT}, ERMVP~\cite{Zhang2024ERMVP}, SICP~\cite{Qu2024SICP}, and mmCooper~\cite{Liu2025mmCooper}) on both simulated (OPV2V) and real-world (DAIR-V2X) datasets, surpassing the second-best methods by 4.10\%/3.25\% and 5.18\%/2.85\% in AP@0.7/0.5, respectively. This superior performance stems from two key factors: first, the DPR module transmits compact codebook indices that are reconstructed into point clouds via a shared codebook, preserving more raw spatial information than other baselines and thus enabling higher detection accuracy; second, the CAF module cascades the ego vehicle’s features using filtered collaborative information before pyramid-scale fusion, further strengthening perception of occluded or uncertain regions and boosting detection precision.\\
\begin{figure*}
  \vspace{-1.0cm} 
  \centering
  \includegraphics[width=1.0\linewidth]{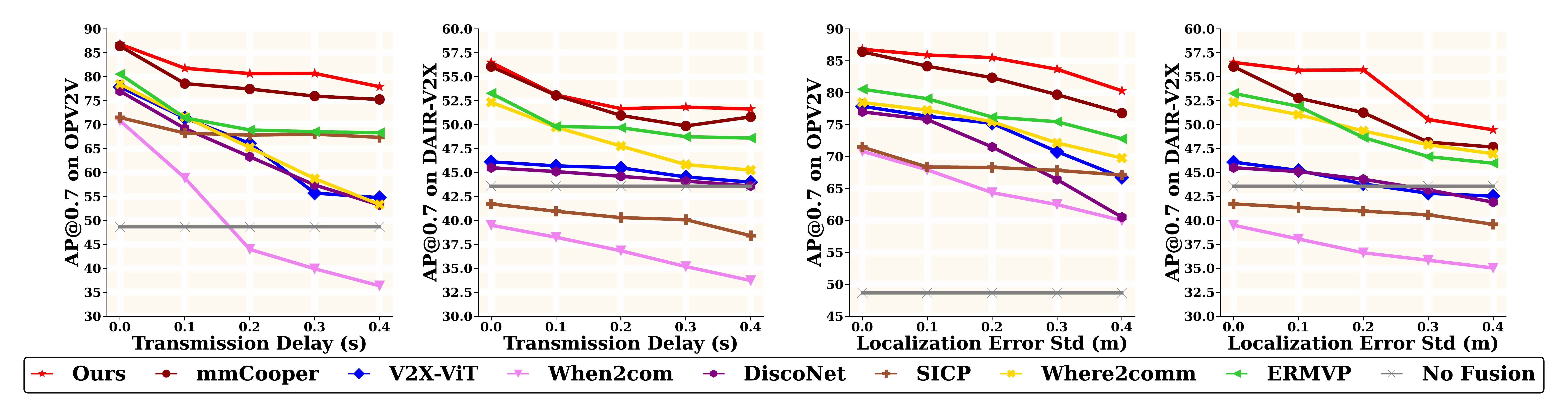}
  \caption{Robustness to the transmission delay and localization error on the OPV2V and DAIR-V2X.}
  \label{fig:robustness}
  \vspace{-0.8cm} 
\end{figure*}

\noindent \textbf{Communication Volume and Transmission Latency.} \cref{fig:main result} shows that QPoint2Comm achieves the best performance with a low and stable communication volume of \textbf{14.95} (log$_2$ scale), which is invariant to scene or traffic variations. This stability arises from the DPR module, which transmits compact discrete codebook indices instead of feature maps, resulting in a fixed and deterministic communication cost.
As shown in \cref{fig:robustness}, perception performance consistently degrades as communication latency increases; thus, lower bandwidth naturally implies lower latency. Our method not only incurs significantly lower bandwidth—leading to much lower latency in practice—but also outperforms other methods even under the same fixed-latency assumption, making the comparison conservative.
 For example, under a hypothetical communication setting with a bandwidth of 1 MB per 100 ms, our method requires only 0.032 MB ($2^{14.95}$ B) per frame, corresponding to 3.3 ms per frame, while Where2comm requires 6.15 MB ($2^{22.62}$ B), i.e., about 615 ms per frame. This highlights that our method would operate at much lower latency in practice.\\
 
\begin{table}[htbp]
\vspace{-1.0cm}
\centering
\caption{Detection Accuracy under Different Packet Loss Rates.}
\vspace{-10pt}
\label{tab:packet_loss}
\resizebox{0.6\textwidth}{!}{ 
\begin{tabular}{c|c|c}  
\hline
\multirow{2}{*}{\textbf{Packet Loss Rate}} & \textbf{OPV2V} & \textbf{DAIR-V2X} \\
\cline{2-3}
                                           & \textbf{AP@0.7 / AP@0.5} & \textbf{AP@0.7 / AP@0.5} \\
\hline
0   & \textbf{82.21} / \textbf{92.18} & \textbf{53.45} / \textbf{67.97} \\
\hline
10\% & 77.78 / 89.06 & 52.09 / 64.32 \\
\hline
20\% & 76.47 / 86.53 & 51.29 / 62.01 \\
\hline
30\% & 75.84 / 85.44 & 50.84 / 60.64 \\
\hline
40\% & 75.67 / 84.93 & 50.53 / 60.00 \\
\hline
\end{tabular}
}
\end{table}
\vspace{-0.4cm}

\noindent \textbf{Robustness to the Transmission Delay and Localization Error.} As shown in \cref{fig:robustness}, our method consistently outperforms all baselines under both temporal and localization perturbations~\cite{Liu2025mmCooper}. Across increasing delays(0-0.4s)  and localization errors(0-0.4m) , performance degradation remains minimal. 
The robustness of QPoint2Comm arises from three key components: 
(1) the DPR transmits compact codebook indices while preserving richer geometric information;
(2) the Mask Training  mechanism trains the network with randomly masked features, improving robustness and enhancing the model’s resistance to perturbations and noise; 
(3) the CAF cascades ego features before fusion using collaborative information, compensating for misalignment caused by both delays and localization errors.
In contrast, When2com~\cite{Liu2020When2com} and SICP~\cite{Qu2024SICP} are less robust to real-world noise and can even underperform the non-fusion baseline, whereas QPoint2Comm remains stable and reliable.\\
\textbf{Ours vs Feature-level-quantization.} Our method differs from feature-level quantization methods (e.g., CodeFilling~\cite{YueCodeFilling:CVPR2024}) in three aspects: we quantize raw LiDAR instead of latent features, reconstruct geometrically structured point clouds rather than feature maps, and transmit voxel-aligned indices with explicit spatial meaning instead of feature-semantic codes.
As shown in \cref{tab:comparison_codefilling_opv2v}, our method outperforms CodeFilling~\cite{YueCodeFilling:CVPR2024} on OPV2V under different latency settings for both AP@0.5 and AP@0.7.

\noindent\textbf{Packet-Loss Tolerance.} We evaluate QPoint2Comm under random packet loss to simulate real-world vehicular communication (\cref{tab:packet_loss}), where transmitted codebook indices are randomly dropped. To address this, Mask Training randomly masks features during training and learns a learnable feature, which is used at inference to fill missing regions caused by packet loss. Together with DPR, which preserves rich geometric information under limited bandwidth, QPoint2Comm remains robust even under severe loss.
Experiments on OPV2V and DAIR-V2X show that, despite gradual degradation, our method consistently outperforms most baselines, demonstrating reliable detection at high packet loss rates.



\subsection{Ablation Study}

\textbf{Effect of Codebook Configuration.} As shown in \cref{tab:codebook_ablation}, detection peaks with a codebook of 2048 and vector dimensionality 1024. Smaller codebooks limit quantization vectors, reducing reconstruction fidelity and degrading detection, while higher-dimensional vectors better capture local geometry but may introduce redundancy. A moderately large codebook with sufficient dimensionality thus balances accurate spatial representation with low bandwidth.\\
\textbf{Importance of Core Components.}
We conduct an ablation study on OPV2V and DAIR-V2X in \cref{tab:ablation_modules} evaluates the contribution of each module. 
Directly using raw point clouds—the only case without DPR—slightly increases accuracy but drastically raises communication cost, demonstrating that our DPR effectively balances accuracy detection with low bandwidth usage.
Excluding the MSK mechanism leads to a clear drop in detection performance, even under ideal transmission conditions, indicating that random masking feature and learnable feature filling not only provide strong tolerance to packet loss but also enhance the model’s intrinsic robustness.
Eliminating the CAF module removes the pre-fusion cascade of ego features, reducing the ego's feature to perceive occluded and uncertain regions and thereby weakening both accuracy and robustness.
The CFF and BFO modules further enhance performance. Although reconstructed collaborative features are informative, they may contain noise. CFF suppresses low-confidence regions to ensure reliable fusion, while BFO filters and refines bounding boxes to improve overall detection accuracy.
Overall, these results show that DPR, MSK, and CAF are central to achieving efficient, robust, and high-performance collaborative perception, while all modules work synergistically to enhance overall system capability.\\
\textbf{Comparison between Quantizing Point Cloud and Feature.}
As shown in \cref{tab:quantization_comparison}, quantizing raw point clouds outperforms feature quantization in detection accuracy. Our method transmits discrete indices and reconstructs structured point clouds at the ego agent, preserving spatial geometry. In contrast, feature quantization applies secondary compression on extracted features, further reducing spatial fidelity and degrading performance. Overall, point cloud quantization provides a better balance between communication cost and accuracy.

\begin{table}[t]
\centering
\begin{minipage}[t]{0.49\textwidth}
\centering
\caption{Ablation Study of Individual Modules on Detection Performance. Modules: Mask Training (MSK), Confidence-based Feature Filter (CFF), Cascade Attention Fusion (CAF), BBox Filter and Offset (BFO).}
\vspace{-10pt}
\label{tab:ablation_modules}
\resizebox{1.0\linewidth}{!}{%
\renewcommand{\arraystretch}{1.5}
\begin{tabular}{c|c|c|c|c|c}
\hline
\textbf{Dataset} & \textbf{MSK} & \textbf{CFF} & \textbf{CAF} & \textbf{BFO} & \textbf{AP@0.7 / AP@0.5} \\
\hline
\multirow{6}{*}{\textbf{OPV2V}} 
& \textcolor{red}{\ding{55}} & \checkmark & \checkmark & \checkmark & 80.52 / 91.79 \\ 
\cline{2-6}
& \checkmark & \textcolor{red}{\ding{55}} & \checkmark & \checkmark & 81.19 / 91.49 \\ 
\cline{2-6}
& \checkmark & \checkmark & \textcolor{red}{\ding{55}} & \checkmark & 79.29 / 90.51 \\ 
\cline{2-6}
& \checkmark & \checkmark & \checkmark & \textcolor{red}{\ding{55}} & 79.15 / 91.61 \\ 
\cline{2-6}
& \checkmark & \checkmark & \checkmark & \checkmark & \textbf{82.21} / \textbf{92.18} \\ 
\cline{2-6}
& \multicolumn{4}{c|}{\cellcolor{gray!20}\textcolor{black}{Raw point cloud transmission}} & \cellcolor{gray!20}83.69 / 93.25 \\
\hline\hline
\multirow{6}{*}{\textbf{DAIR-V2X}} 
& \textcolor{red}{\ding{55}} & \checkmark & \checkmark & \checkmark & 52.60 / 67.80 \\ 
\cline{2-6}
& \checkmark & \textcolor{red}{\ding{55}} & \checkmark & \checkmark & 50.19 / 66.22 \\ 
\cline{2-6}
& \checkmark & \checkmark & \textcolor{red}{\ding{55}} & \checkmark & 49.13 / 64.91 \\ 
\cline{2-6}
& \checkmark & \checkmark & \checkmark & \textcolor{red}{\ding{55}} & 51.70 / 66.76 \\ 
\cline{2-6}
& \checkmark & \checkmark & \checkmark & \checkmark & \textbf{53.45} / \textbf{67.97} \\ 
\cline{2-6}
& \multicolumn{4}{c|}{\cellcolor{gray!20}\textcolor{black}{Raw point cloud transmission}} & \cellcolor{gray!20}53.59 / 68.21 \\
\hline
\end{tabular}%
}
\end{minipage}
\hfill 
\begin{minipage}[t]{0.49\textwidth}
\centering
\caption{Effect of Codebook Configuration on Detection Performance.}
\vspace{-10pt}
\label{tab:codebook_ablation}
\resizebox{1.0\linewidth}{!}{%
\renewcommand{\arraystretch}{2.2}
\begin{tabular}{c|c|c|c|c}
\hline
\textbf{Dataset} & \textbf{Codebook Size} & \textbf{Code Dim} & \textbf{AP@0.7 / AP@0.5} & \textbf{Communication Volume} \\
\hline
\multirow{4}{*}{\textbf{OPV2V}} 
& 2048 & 1024 & \textbf{82.21} / \textbf{92.18} & 14.95 \\ \cline{2-5}
& 1024 & 1024 & 81.43 / 91.64 & 14.81 \\ \cline{2-5}
& 512  & 1024 & 81.33 / 91.33 & 14.66 \\ \cline{2-5}
& 2048 & 512  & 81.92 / 92.09 & 14.95 \\ \hline
\multirow{4}{*}{\textbf{DAIR-V2X}} 
& 2048 & 1024 & \textbf{53.45} / \textbf{67.97} & 14.95 \\ \cline{2-5}
& 1024 & 1024 & 53.21 / 66.67 & 14.81 \\ \cline{2-5}
& 512  & 1024 & 53.18 / 64.60 & 14.66 \\ \cline{2-5}
& 2048 & 512  & 52.44 / 62.18 & 14.95 \\ \hline
\end{tabular}%
}
\vspace{0.23cm} 

\centering
\caption{Comparison between Quantized Point Cloud and Quantized Feature.}
\vspace{-10pt}
\label{tab:quantization_comparison}
\resizebox{\textwidth}{!}{%
\renewcommand{\arraystretch}{2.0}
\begin{tabular}{c|c|c}
\hline
\multirow{2}{*}{\textbf{Dataset}} & \textbf{Quantized Point Cloud $x_c$}  & \textbf{Quantized Feature $F_c$} \\ 
\cline{2-3}
 & \textbf{AP@0.7 / AP@0.5} & \textbf{AP@0.7 / AP@0.5} \\ 
\hline
\textbf{OPV2V} & \textbf{82.21 / 92.18} & 77.43 / 90.23 \\ \hline
\textbf{DAIR-V2X} & \textbf{53.45 / 67.97} & 49.06 / 66.03 \\ 
\hline
\end{tabular}%
}
\end{minipage}
\vspace{-0.40cm} 
\end{table}

\begin{figure*}
  \centering
  \begin{minipage}[t]{0.48\linewidth} 
    \centering
    \includegraphics[width=1.0\linewidth]{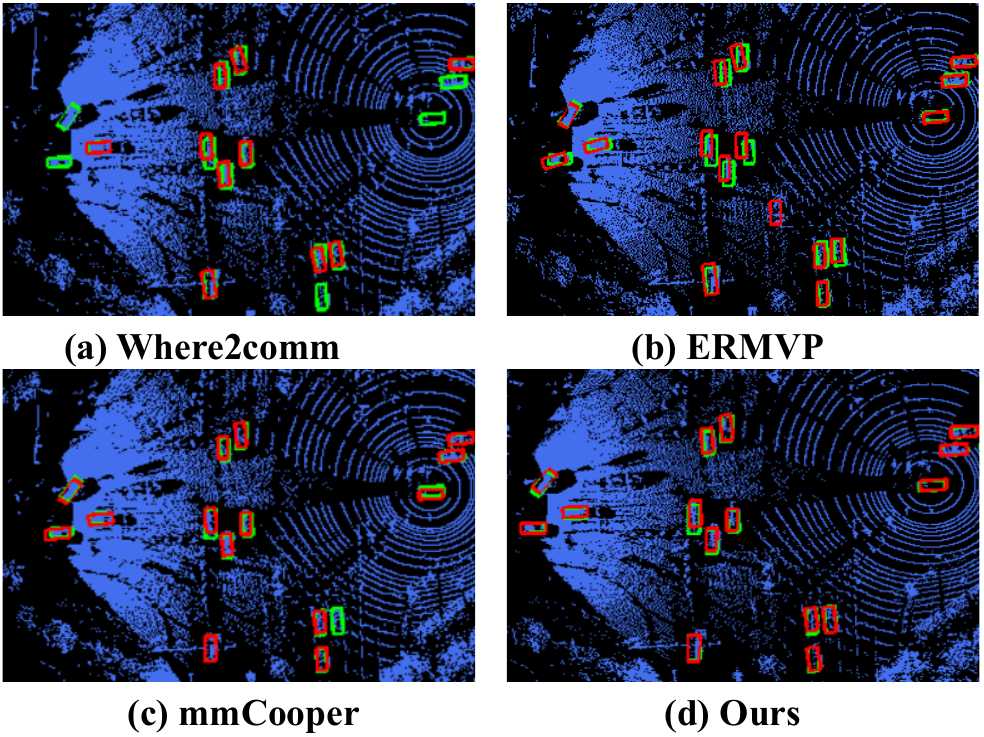}
    \caption{Detection Results on DAIR-V2X (Green: GT, Red: Prediction).}
    \label{fig:visual937}
  \end{minipage}
  \hfill 
  \begin{minipage}[t]{0.48\linewidth} 
    \centering
    \includegraphics[width=1.0\linewidth]{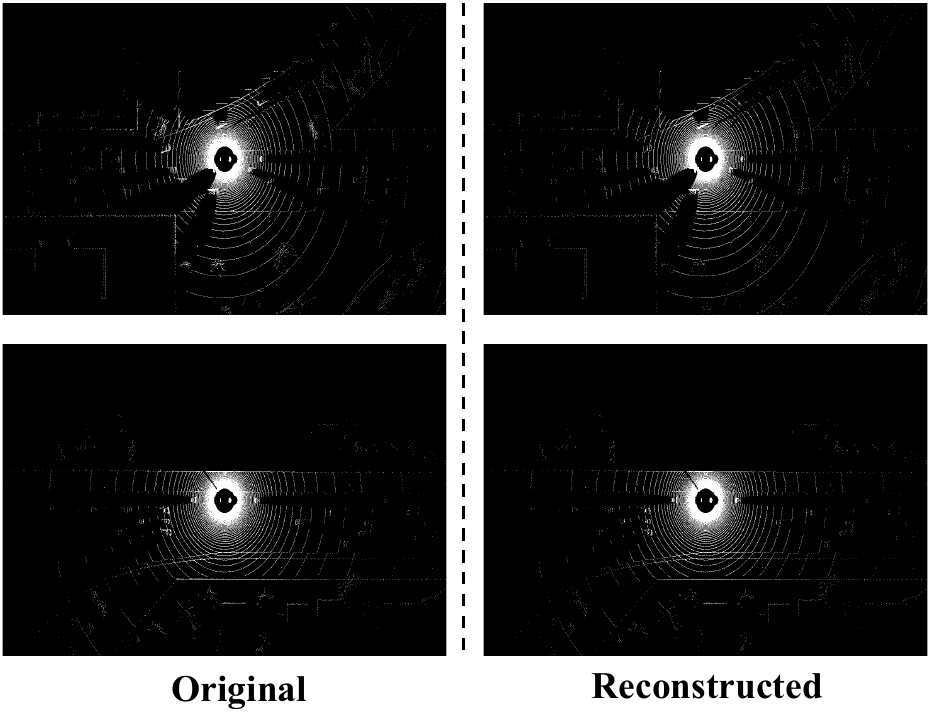}
    \caption{BEV Representations on OPV2V Dataset (Original vs Reconstructed).}
    \label{fig:opv2v_original_reconstruct}
  \end{minipage}
  \vspace{-0.4cm}
\end{figure*}

\subsection{Qualitative Results of Detection Results}
As shown in \cref{fig:visual937}, our model detects more objects in DAIR-V2X scenes, while baselines often miss or misalign targets. This gain comes from DPR preserving richer spatial information and CAF reinforcing ego features with filtered collaborative cues, improving robustness  under challenging conditions.
\subsection{Qualitative Results of Original and Reconstructed Point Clouds}
We evaluate the Discrete Point Cloud Representation (DPR) module by comparing original and reconstructed point clouds on OPV2V (Fig.~\ref{fig:opv2v_original_reconstruct}). The reconstructed point clouds preserve spatial structure and key object information, showing that DPR supports compact transmission with minimal loss for robust collaborative perception.

\section{Conclusion}

In this paper, we present QPoint2Comm, the first collaborative perception framework that employs the Discrete Point Cloud Representation module to quantize point clouds into compact discrete indices via shared codebooks for transmission, achieving low bandwidth while retaining rich raw information.
Combined with the Mask Training, it is also the first framework to provide inherent tolerance to packet loss, ensuring stable perception even under high-loss scenarios.
Furthermore, the Pyramid-scale with Cascade Attention Fusion cascades ego feature before fusion, improving robustness to occlusion and uncertainty. Experiments on OPV2V and DAIR-V2X show that QPoint2Comm achieves superior accuracy with minimal bandwidth and remains robust under severe packet loss, validating its effectiveness and stability.


%
%
\bibliographystyle{splncs04}
\bibliography{main}
\clearpage
\setcounter{page}{1}

\section{Overview of supplementary material}
The supplementary material is organized into the following sections:

\begin{enumerate}
    \item \cref{sec:system pipline}: The Overall System Pipeline
    \begin{enumerate}
        \item \cref{sub:DPR pipline}: The Training Pipeline of Discrete Point Cloud Representation 
        \item \cref{sub:Detection pipline}: The Training  Pipeline of the Mask Fusion and Detection
        \item \cref{sub:inference pipline}: The Inference Pipeline of Our Proposed  QPoint2Comm
    \end{enumerate}
    \item \cref{sec:addtional}: The Additional Experimental Results on OPV2V and DAIR-V2X Datasets
    \begin{enumerate}
        \item \cref{sub:implement details}: Implementation Details
        \item \cref{sub:supplement experiments}: Supplements on Localization Errors, Transmission Delays and Heading Errors
       \item \cref{sub:ablation backbone}: Ablation Study on Separated Backbone and Shared Backbone
    \end{enumerate}

    \item \cref{sec:qualitative}: The Additional Qualitative Results on Point Cloud Reconstruction and Detection Results
    \begin{enumerate}
        \item \cref{sub:visual DPR}: Visualization of Point Cloud Reconstruction
        \item \cref{sub:visual Detection}: Visualization of Detection Results
    \end{enumerate}
    \item \cref{chamfer_distance}: Quantitative comparsion of original and reconstructed point cloud
    \item  \cref{feature_vs_ours}: Feature-level quantization vs Ours
    \item  \cref{more_settings}: More quantitave results under different settings    
\end{enumerate}

\section{The System Pipeline of our models}
\label{sec:system pipline}
Our model is trained in two stages: the first stage learns the discrete representation of point cloud (\cref{sub:DPR pipline}), and the second stage focuses on mask training, feature fusion, and bounding-box generation (\cref{sub:Detection pipline}). After these two stages, we further describe the inference pipeline of the QPoint2Comm system (\cref{sub:inference pipline}).

\subsection{The Training Pipeline of Discrete Point Cloud Representation Module}
\label{sub:DPR pipline}
The training pipeline of the \textbf{Discrete Point Cloud Representation (DPR)} module is illustrated in \cref{alg:dpr_pretrain}. The algorithm learns discrete representations of LiDAR point cloud by jointly training the occupancy and intensity branches, enabling the model to capture complementary geometric and reflectance characteristics and providing structured encodings for downstream collaborative perception tasks.

\textbf{Grid-based Preprocessing:} Grid-based preprocessing converts the raw LiDAR point cloud \({x}\) into two independent tensors: an occupancy tensor \({X}_{\text{occ}}\in\{0,1\}^{H\times W\times L}\), indicating whether each grid contains points, and an intensity tensor \({X}_{\text{int}}\in[0,1]^{H\times W\times L}\), representing normalized reflectivity. This discretization captures the spatial structure of the point cloud. 

\textbf{Codebook Training:} Branch-specific training is performed, with occupancy and intensity learned independently. In the occupancy branch, the encoder \(E_{\text{occ}}\) maps \({X}_{\text{occ}}\) to latent features \({Z}_{\text{occ}}\), which are then quantized using a learnable codebook \(\mathcal{E}_{\text{occ}}\) to produce discrete embeddings \(\hat{{Z}}_{\text{occ}}\). A decoder \(\text{Dec}_{\text{occ}}\), combined with grid-wise Gaussian parameters, reconstructs the occupancy tensor. The training objective includes a vector quantization loss \(\mathcal{L}_{\text{vq,occ}}\) and a reconstruction loss \(\mathcal{L}_{\text{occ}}\) based on binary cross-entropy. 
The intensity branch uses a similar encoder–codebook–decoder pipeline 
($E_{\text{int}}, \mathcal{E}_{\text{int}}, \text{Dec}_{\text{int}}$) 
with supervision via mean squared error ($\mathcal{L}_{\text{int}}$) 
and quantization loss ($\mathcal{L}_{\text{vq,int}}$). 
Unlike occupancy, the intensity decoder does not apply Gaussian sampling, 
producing the same reflection value for all points within a given occupied grid.
The codebooks for the occupancy and intensity branches are updated via K-means clustering to adaptively capture their respective distributions. The framework maintains separate encoders, decoders, and codebooks for occupancy and intensity, producing structured, type-specific discrete representations. This design enables efficient discrete representations for cross-agent transmission and collaborative perception.

\begin{algorithm*}[!htbp]
\caption{The Training Pipeline of the Discrete Point Cloud Representation }
\label{alg:dpr_pretrain}
\KwIn{Raw LiDAR point cloud from collaborative agents $\{{x}_c\}^{N_c}.$}
\KwOut{Trained occupancy encoder $E_{\text{occ}}$, decoder $\text{Dec}_{\text{occ}}$, codebook $\mathcal{E}_{\text{occ}}=\{\mathbf{e}_k^{\text{occ}}\}_{k=1}^{K_{\text{occ}}}$;\\
\qquad Intensity encoder $E_{\text{int}}$, decoder $\text{Dec}_{\text{int}}$, codebook $\mathcal{E}_{\text{int}}=\{\mathbf{e}_k^{\text{int}}\}_{k=1}^{K_{\text{int}}}.$}
\smallskip

\textbf{1. Voxelization:} Discretize ${x_c}$ into two independent tensors:
\begin{itemize}
    \item Occupancy tensor ${X}_{\text{occ}}\!\in\!\{0,1\}^{\mathcal{H}\!\times\!\mathcal{W}\!\times\!\mathcal{L}};$
    \item Intensity tensor ${X}_{\text{int}}\!\in\![0,1]^{\mathcal{H}\!\times\!\mathcal{W}\!\times\!\mathcal{L}}.$
\end{itemize}

\textbf{2. Separate Training for Each Branch:}
\For{branch $\in \{\text{occupancy}, \text{intensity}\}$}{
    \eIf{branch is occupancy}{
        \textbf{2.1 Encoding:} Obtain latent features ${Z}_{\text{occ}} = E_{\text{occ}}({X}_{\text{occ}})\!\in\!\mathbb{R}^{h\times w\times D_{\text{occ}}}.$\\[2pt]
        
        \textbf{2.2 Vector Quantization:} For each ${z}_{i}^{\text{occ}}\!\in\!{Z}_{\text{occ}}$, find nearest codebook entry:
        \[
        k_{\text{occ}}^* = \arg\min_k \|{z}_{i}^{\text{occ}} - {e}_k^{\text{occ}}\|_2^2,\quad 
        \hat{{Z}}_{\text{occ}} = q_{\text{occ}}({Z}_{\text{occ}}) = \{{e}_{k_{\text{occ}}^*}^{\text{occ}}\}.
        \]\\[-3pt]
        
        \textbf{2.3 Reconstruction:} Reconstruct occupancy tensor:
        \[
        \hat{{X}}_{\text{occ}} = \text{Dec}_{\text{occ}}(\hat{{Z}}_{\text{occ}}; \mathcal{N}({\mu}_{\text{occ}}, \Sigma_{\text{occ}})).
        \]
        
        \textbf{2.4 Loss Computation:}
        \[
        \mathcal{L}_{\text{occ}} = \mathcal{L}_{\text{vq,occ}} + \lambda_{\text{occ}} \mathcal{L}_{\text{occ\_reconstruct}}.
        \]
        
        \textbf{2.5 Codebook Update:} Update $\mathcal{E}_{\text{occ}}$ via exponential moving average.\\[2pt]
    }{
        \textbf{2.6 Encoding:} Obtain latent features ${Z}_{\text{int}} = E_{\text{int}}({X}_{\text{int}})\!\in\!\mathbb{R}^{h\times w\times D_{\text{int}}}.$\\[2pt]
        
        \textbf{2.7 Vector Quantization:} For each ${z}_{i}^{\text{int}}\!\in\!\mathbf{Z}_{\text{int}}$, find nearest codebook entry:
        \[
        k_{\text{int}}^* = \arg\min_k \|{z}_{i}^{\text{int}} - \mathbf{e}_k^{\text{int}}\|_2^2,\quad 
        \hat{{Z}}_{\text{int}} = q_{\text{int}}({Z}_{\text{int}}) = \{{e}_{k_{\text{int}}^*}^{\text{int}}\}.
        \]\\[-3pt]
        
        \textbf{2.8 Reconstruction:} Reconstruct intensity tensor:
        \[
        \hat{{X}}_{\text{int}} = \text{Dec}_{\text{int}}(\hat{{Z}}_{\text{int}}).
        \]        
        \textbf{2.9 Loss Computation:}
        \[
        \mathcal{L}_{\text{int}} = \mathcal{L}_{\text{vq,int}} + \lambda_{\text{int}} \mathcal{L}_{\text{int\_reconstruct}}.
        \] 
        
        \textbf{2.10 Codebook Update:} Update $\mathcal{E}_{\text{int}}$ via exponential moving average.\\[2pt]
    }
}

\textbf{3. Output:} Store all trained components $\{E_{\text{occ}}, \text{Dec}_{\text{occ}}, \mathcal{E}_{\text{occ}}\}$ and $\{E_{\text{int}}, \text{Dec}_{\text{int}}, \mathcal{E}_{\text{int}}\}$ as $\Theta_{\text{DPR}}.$
\end{algorithm*}

\subsection{The Training  Pipeline of the Mask Fusion and Detection Module}
\label{sub:Detection pipline}

\noindent
As shown in \cref{alg:detection_pipeline}, the \textbf{Mask Fusion and Detection} module is trained using the raw ego point cloud $x_e$, collaborative point clouds$\{x_c\}^{N_c}$, and ground-truth bounding boxes $B_{\text{gt}}$, and produces the trained detection parameters $\Theta_{\text{det}}$. 

\textbf{Mask Training:} Ego features $F_e$ are first extracted from $x_e$. Each collaborative point cloud $x_c$ is processed to obtain features $\tilde{F}$, where random regions are masked and replaced with learnable features $f_{\text{learnable}}$ to produce the filled collaborative features $F_c$.

\textbf{Feature Fusion and Detection:} 
The collaborative features $F_c$ are first filtered using a Confidence-based Feature Filter module (CFF) to produce $\hat{F}_c$ and $\hat{B}_c$. Then, cascade ego feature $F_e$ with filtered collaborative features $\hat{F}_c$ to obtain cascaded ego feature $\hat{F}_e$. Pyramid-scale fusion aggregates information across different feature scales to obtain fused features $\mathcal{F}_i$, which are then fed into the detection head to predict fused bounding boxes $B_{\text{fused}}$. Collaborative bounding boxes $\hat{B}_c$ is corrected using the BBox Filter and Offset (BFO) module to produce $B_{\text{offset}}$. The final bounding boxes $B_{\text{final}}$ are obtained by merging $B_{\text{fused}}$ and $B_{\text{offset}}$.

\textbf{Loss Computation:} 
The predicted bounding boxes $B_{\text{final}}$ are compared against the ground-truth $B_{\text{gt}}$ to compute the total loss $L_{\text{total}}$, which includes regression, classification, offset, and confidence score terms. This loss is used to optimize the detection model parameters $\Theta_{det}$.

\begin{algorithm*}[!htbp]
\caption{The Training Pipeline of the Mask Fusion and Detection}
\label{alg:detection_pipeline}
\KwIn{Raw ego LiDAR point cloud $x_e$, raw collaborative point clouds $\{x_c\}^{N_c}$, and ground-truth bounding boxes $B_{\text{gt}}.$}
\KwOut{Trained detection model parameters $\Theta_{det}.$}
\smallskip

\textbf{1. Mask Training:}

\textbf{1.1 Ego Feature Extraction:}
\[
F_e = \psi_E(x_e).
\]

\textbf{1.2 Collaborative Feature Masking and Filling:}\\
\For{each collaborative point cloud $x_c$}{
    Extract features masking:$\tilde{F} =\mathcal{M}_r\big(\psi_C(x_c)\big).$
    
   Apply learnable feature filling: 
   
   Masked features in $\tilde{F}$ are replaced with the learnable feature $f_{\text{learnable}}$, producing $F_c$.

}

\smallskip
\textbf{2. Feature Fusion and Detection:}

\textbf{2.1 Confidence-based Feature Filter:}\\
\For{each filled collaborative feature $F_c$}{
\[
B_c = \text{Detector}(F_c), \quad 
\hat{G}_{f,b} = G_{f,b} \odot \mathbb{I}_{\{G_{f,b} > \tau_p\}} \odot \mathcal{G}_\sigma(G_{f,b}), \quad\]
\[\hat{F}_c = \hat{G}_f \otimes F_c, \quad 
\hat{B}_c = \hat{G}_b \otimes B_c.
\]

}

\textbf{2.2 Ego Feature Cascade:}\\
Cascade ego feature with collaborative features:
\[
A_{\text{add}} = H_{\text{add}}(\text{Attn}(F_e, \hat{F}_c, \hat{F}_c)) + F_e, \quad
A_{\text{mul}} = H_{\text{mul}}(\text{Attn}(F_e, \hat{F}_c, \hat{F}_c)) \otimes F_e, \quad\]
\[\hat{F}_e = \sigma(W_g) \otimes A_{\text{add}} + (1-\sigma(W_g)) \otimes A_{\text{mul}}.
\]

\textbf{2.3 Pyramid-scale Feature Fusion:}\\
\For{each filtered collaborative feature $\hat{F}_c$}{
\[
\mathcal{F}_i^l = \text{NeighborCrossAttn}(MLP(\hat{F}_e^l), \hat{F}_c^l, \hat{F}_c^l), \, l=1,2,3, \quad
\mathcal{F}_i = \text{Concat}(\mathcal{F}_i^1, \mathcal{F}_i^2, \mathcal{F}_i^3)
.\]
}

\textbf{2.4 Detection Head:}\\
Predict bounding boxes:
\[
B_{\text{fused}} = \text{Detector}(\mathcal{F}_i).
\]

\textbf{2.5 BBox Filter and Offset:}\\
\[
(\text{off}, \text{score}) = \text{Deformable-Bbx-Attention}(\mathcal{F}_i, \hat{B}_c), \quad
B_{\text{offset}} = \phi(\hat{B}_c, \text{off}, \text{score}).
\]

\textbf{2.6 Final Fusion:}
\[
B_{\text{final}} = \text{Merge}(B_{\text{fused}}, B_{\text{offset}}).
\]

\smallskip

\textbf{3. Loss Computation:}\\
Compute the training loss between the final predictions $B_{\text{final}}$ and ground-truth $B_{\text{gt}}$:
\[
L_{\text{total}} = L_{\text{reg}} + L_{\text{cls}} + L_{\text{off}} + L_{\text{score}}.
\]
\smallskip
\textbf{Output:} Return trained detection model parameters $\Theta_{det}$.
\end{algorithm*}

\subsection{The Inference  Pipeline of the Qpoint2Comm}
\label{sub:inference pipline}
\noindent
As illustrated in \cref{alg:inference_pipeline}, the inference pipeline of QPoint2Comm takes as input the raw ego LiDAR point cloud $x_e$, raw collaborative point clouds $\{x_c\}^{N_c}$, and the trained parameters of the discrete point cloud representation (DPR) $\Theta_{\text{DPR}}$ and detection module $\Theta_{\text{det}}$. The output is the final bounding boxes $B_{\text{final}}$. 

\textbf{Discrete Point Cloud Representation:} 
Each collaborative point cloud $x_c$ is first encoded into discrete codebook indices using the DPR module. These indices are transmitted under lossy network conditions and subsequently decoded to reconstruct the point cloud $\hat{x}_c$ in the ego vehicle's coordinate frame. This process compresses the collaborative data for efficient transmission while preserving the essential spatial and intensity information necessary for downstream perception tasks.

\textbf{Fusion and Detection:}  
First, ego features $F_e$ are extracted from the ego point cloud $x_e$. Each reconstructed collaborative point cloud $\hat{x}_c$ is processed to extract features $\tilde{F}$, where missing regions are filled using learnable feature $f_{\text{learnable}}$ to produce $F_c$. The collaborative features $F_c$ and corresponding bounding boxes $B_c$ are filtered with the CFF module to obtain filtered features $\hat{F}_c$ and bounding boxes $\hat{B}_c$. Then, cascade ego feature $F_e$ with filtered collaborative features $\hat{F}_c$ to obtain cascaded ego feature $\hat{F}_e$. Pyramid-scale fusion combines $\hat{F}_e$ with $\hat{F}_c$ across different scales to obtain fused features $\mathcal{F}_i$. The fused features are fed into the detection head to predict bounding boxes $B_{\text{fused}}$. In parallel, collaborative bounding boxes $\hat{B}_c$ are refined using the BFO module to produce $B_{\text{offset}}$. Finally, the fused and refined bounding boxes are merged to produce the final predictions $B_{\text{final}}$.

This pipeline effectively integrates the ego vehicle’s information with reconstructed collaborative point clouds, compensates for missing or occluded regions in the collaborative data, and produces robust and accurate 3D object detections, ensuring reliable perception even under lossy communication conditions.

\begin{algorithm*}[t]
\caption{Inference Pipeline of QPoint2Comm}
\label{alg:inference_pipeline}
\KwIn{
Raw ego LiDAR point cloud $x_e$, raw collaborative point clouds $\{x_c\}^{N_c}$, 
trained DPR parameters $\Theta_{\text{DPR}}$, trained detection parameters $\Theta_{\text{det}}.$
}
\KwOut{Final fused bounding boxes $B_{\text{final}}.$}
\smallskip

\textbf{1. Discrete Point Cloud Representation:}

\For{each collaborative point cloud $x_c$}{
Encode $x_c$  into discrete indices, transmit in lossy condition, and decode to obtain reconstructed point cloud $\hat{x_c}$.
}

\smallskip
\textbf{2.Fusion and Detection:}

\textbf{2.1 Ego Feature Extraction:} $F_e = \psi_E(x_e).$

\textbf{2.2 Learnable Feature Filling:}

Extract features:
$\tilde{F} = \psi_C(\hat{x}_c).$

Let $\mathcal{S}$ denote the set of feature locations that are missing due to packet loss.

\For{each feature location $j$}{
    \eIf{$j \in \mathcal{S}$}{
        Fill the missing region with the learnable representation:  
        $F_c[j] = f_{\text{learnable}}$,
    
    }{
        Use the reconstructed collaborative feature:  
        $F_c[j] = \tilde{F}[j]$.
        
    }
}

\textbf{2.3 Confidence-based Feature Filter:} 
Filter unreliable regions in collaborative features $F_c$ to get $\hat{F}_c$ and filter collaborative bounding boxes ${B}_c$ to get $\hat{B}_c$ .

\textbf{2.4 Ego Feature cascade :} Cascade ego feature $F_e$ with $\hat{F}_c$ to obtain cascaded ego feature $\hat{F}_e$.

\textbf{2.5 Pyramid-scale Collaborative Fusion:} Fuse cascaded ego feature $\hat{F}_e$ with $\hat{F}_c$ across multiple-scale to get fused features $\mathcal{F}_i$.

\textbf{2.6 Detection Head:} Fused bounding boxes, $B_{\text{fused}} = \text{Detector}(\mathcal{F}_i)$.

\textbf{2.7 BBox Filter and Offset:} Refine collaborative bounding boxes  $\hat{B}_c$ using BFO module to obtain $B_{\text{offset}}$.

\textbf{2.8 Final Merge:} Merge fused and refined boxes to produce final predictions: 
$B_{\text{final}} = \text{Merge}(B_{\text{fused}}, B_{\text{offset}})$.

\smallskip
\textbf{Output:} Return final bounding boxes $B_{\text{final}}$.
\end{algorithm*}

\section{Additional Experimental Results on OPV2V and DAIR-V2X}
\label{sec:addtional}

\subsection{Implementation Details}
\label{sub:implement details}
\textbf{DPR Module.}
On the OPV2V~\cite{Xu2022OPV2V} and DAIR-V2X~\cite{Yu2022DAIRV2X} datasets, the grid size is set to $0.15625 \times 0.15625 \times 0.15$ meters along the $x$, $y$, and $z$. 
The latent representation has a spatial resolution of $h \times w = 80 \times 1440$ and a channel dimension of $D = 1024$, resulting in $h \times w = 11520$ vectors of dimension $D$ to be quantized.
The vector quantizer maintains reliable codebook usage through a reservoir-based refresh and periodic K-means reinitialization when code frequency falls below the dead limit (256), ensuring stable and effective quantization throughout training.\\  
\textbf{Dection Module.}
On the OPV2V~\cite{Xu2022OPV2V} and DAIR-V2X~\cite{Yu2022DAIRV2X} datasets, the voxel grid size encoded by the backbone encoder is set to $0.4 \times 0.4 \times 4$ meters along the $x$, $y$, and $z$. 
The resulting bird’s-eye-view (BEV) feature maps, shared among collaborative agents, have channel dimension $C=64$ and spatial dimensions $H \times W = 100 \times 352$ for OPV2V, and $H \times W = 100 \times 252$ for DAIR-V2X.
Shared bounding boxes among agents are parameterized by their center coordinates, object dimensions (length, width, height), and heading angle. 
The detection head is implemented using two separate $1\times1$ convolutional layers, one for regression of geometric parameters and the other for classification of object confidence scores.

\subsection{Supplements on Localization Errors, Transmission Delays and Heading Errors}
\label{sub:supplement experiments}

\begin{figure*}[!htbp]
  \centering
  \includegraphics[width=1.0\linewidth]{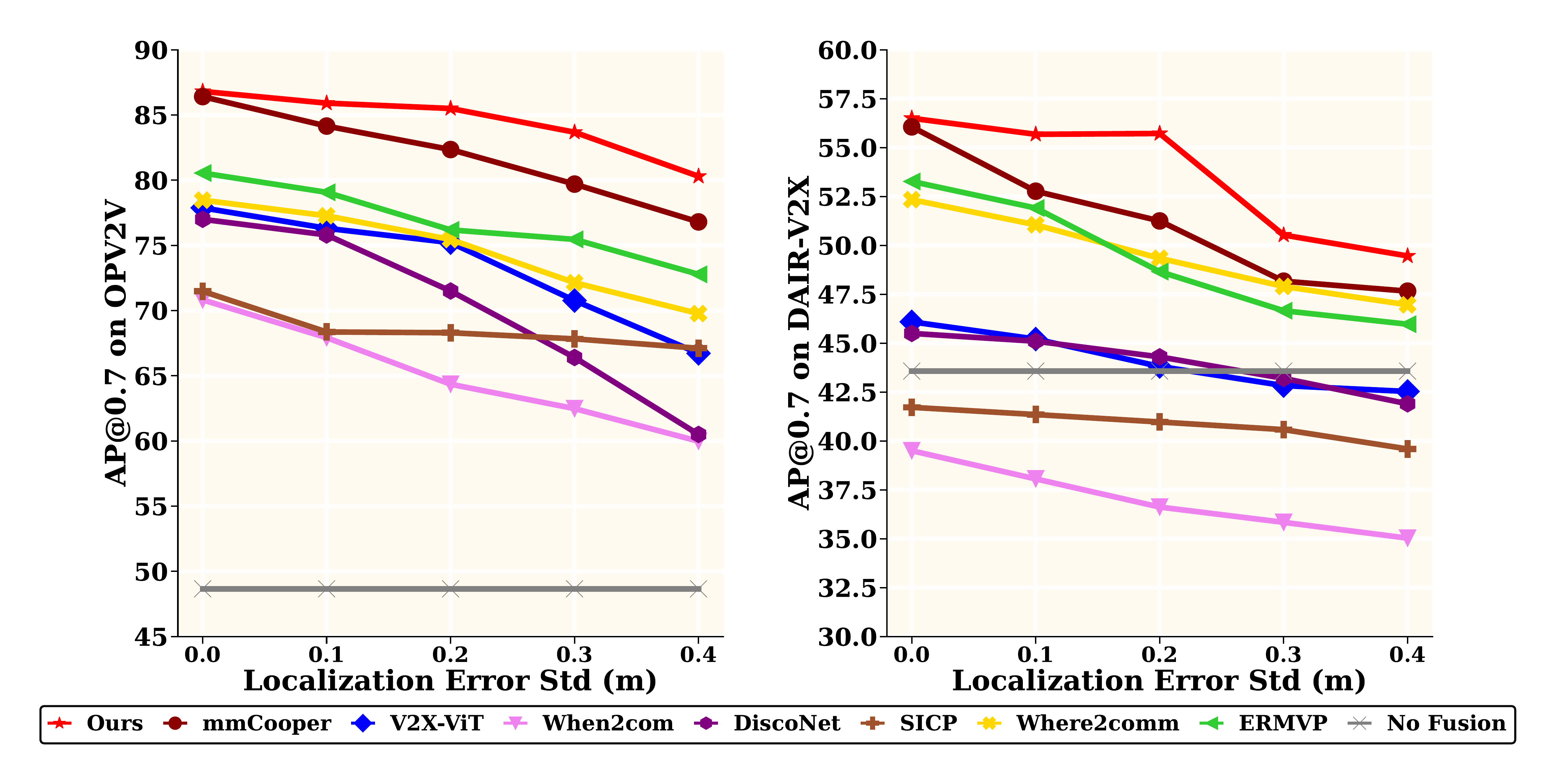}
  \caption{Robustness to the localization error on the OPV2V and DAIR-V2X datasets.}
  \label{fig:loc error}
\end{figure*}

\begin{figure*}[!htbp]
  \centering
  \includegraphics[width=1.0\linewidth]{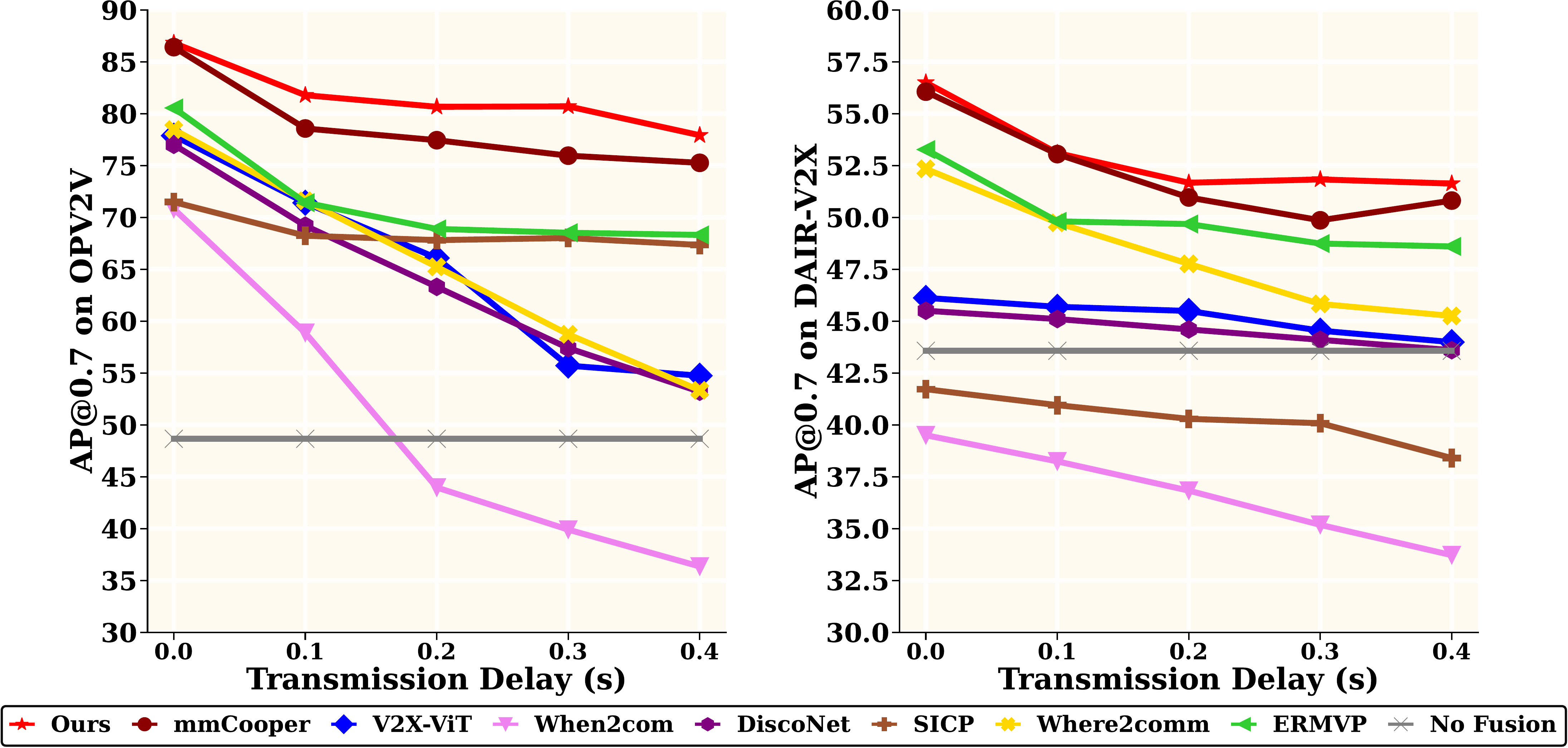}
  \caption{Robustness to the transmission delay on the OPV2V and DAIR-V2X datasets.}
  \label{fig:time delay}
\end{figure*}

\begin{figure*}[!htbp]
  \centering
  \includegraphics[width=1.0\linewidth]{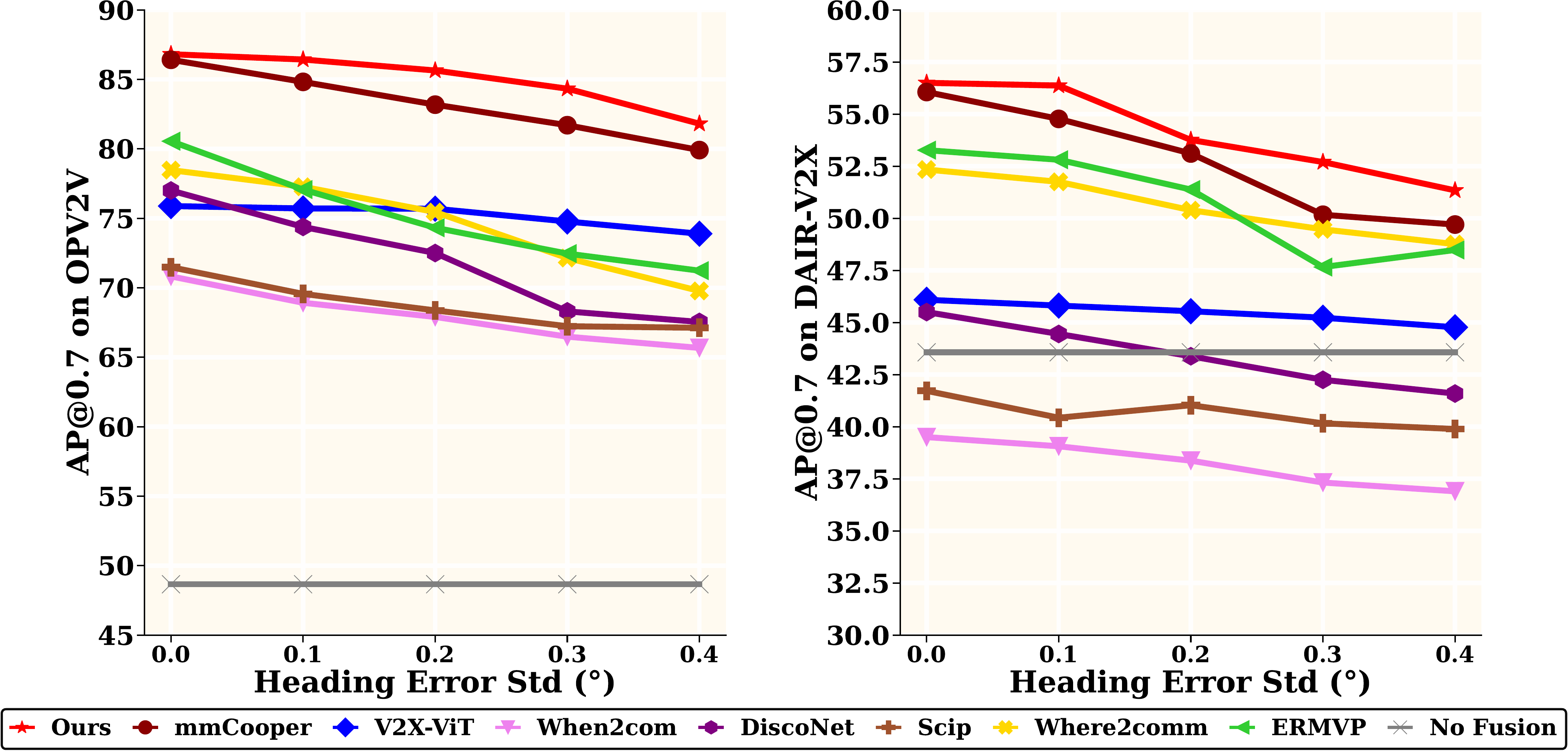}
  \caption{Robustness to the heading error on the OPV2V and DAIR-V2X datasets.}
  \label{fig:heading error}
\end{figure*}

We evaluate the robustness of the proposed method on the OPV2V and DAIR-V2X datasets under three types of perturbations: localization errors, transmission delays, and heading noise. Localization errors are sampled from a Gaussian distribution with zero mean and standard deviation $\sigma \in \{0.0, 0.1, 0.2, 0.3, 0.4\}$ m, and the results in \cref{fig:loc error} indicate that the method consistently outperforms existing state-of-the-art approaches~\cite{Liu2025mmCooper,Xu2022V2XViT,Liu2020When2com,Qu2024SICP,Hu2022Where2comm,Zhang2024ERMVP,Xu2022OPV2V,Li2021DistilledGraph}, across all levels of error. Similarly, when evaluating the impact of transmission delays of $\{0,100,200,300,400\}$ ms, \cref{fig:time delay} shows that the proposed approach maintains superior performance compared to baseline methods under all delay conditions. Finally, as shown in \cref{fig:heading error}, the method demonstrates robustness to heading errors ranging from $\{0.0, 0.1, 0.2, 0.3, 0.4\}^\circ$, consistently outperforming other models despite the gradual decrease in detection accuracy with increasing noise. Collectively, these results highlight the robustness of the proposed method to common real-world perturbations in collaborative perception scenarios.

\subsection{Ablation Study on Separated Backbone and Shared Backbone}
We conducted experiments on OPV2V and DAIR-V2X using a PointPillars-based~\cite{Lang2019PointPillars} backbone, comparing separate encoders with a shared encoder, as shown in \cref{tab:backbone_comparison}. 
In the separated-backbone setting, the ego LiDAR point cloud and the reconstructed collaborative point clouds are processed by two independent backbone encoders, allowing each encoder to better model the distinct feature characteristics of ego point cloud and collaborative point clouds. 
By contrast, a shared encoder must handle both inputs with the same set of parameters, making it difficult to capture their different feature distributions. 
As evidenced by the experimental results, the separated-backbone configuration consistently achieves higher detection accuracy, demonstrating its superiority over the shared-backbone alternative.

\label{sub:ablation backbone}
\begin{table}[H]
\centering
\caption{Comparison between Separated Backbone and Shared Backbone. Metric: AP@0.7 / AP@0.5.}
\label{tab:backbone_comparison}
\resizebox{0.8\textwidth}{!}{%
\begin{tabular}{c|c|c}
\hline
\multirow{2}{*}{\textbf{Dataset}} & \textbf{Separated Backbone }  & \textbf{Shared Backbone} \\ 
\cline{2-3}
 & \textbf{AP@0.7 / AP@0.5} & \textbf{AP@0.7 / AP@0.5} \\ 
\hline
\multirow{1}{*}{\textbf{OPV2V}} & \textbf{82.21 / 92.18} & 79.87 / 91.60 \\ \hline
\multirow{1}{*}{\textbf{DAIR-V2X}} & \textbf{53.45 / 67.97} & 50.36 / 65.55 \\ 
\hline
\end{tabular}%
}
\end{table}

\section{Additional Qualitative Results}
\label{sec:qualitative}

\subsection{Visualization of Original and Reconstructed Point Clouds}
\label{sub:visual DPR}

\begin{figure*}[!htbp]
  \centering
  \includegraphics[width=1.0\linewidth]{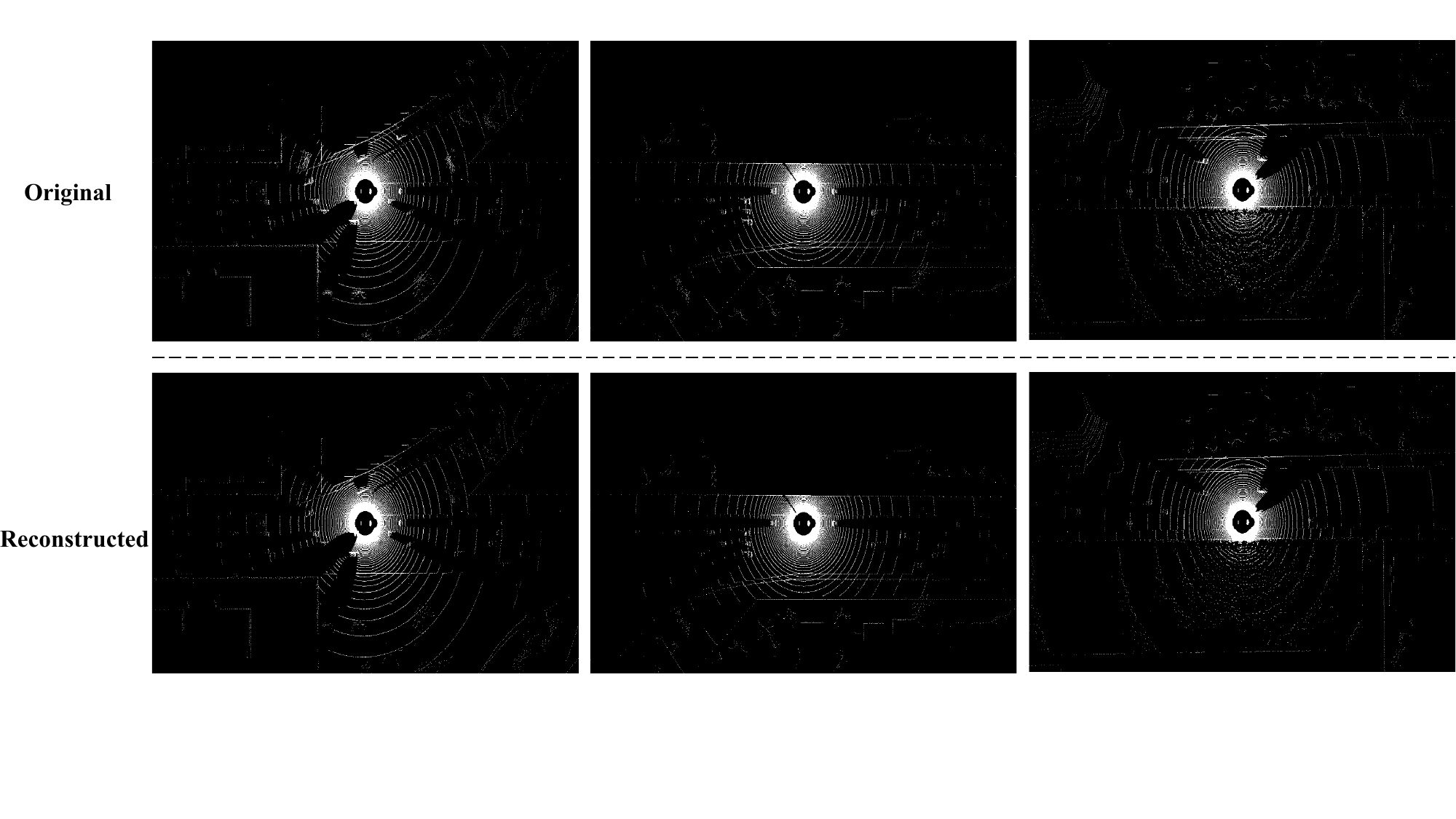}
  \caption{Visualization of BEV representations on the OPV2V dataset. 
    The top row shows the original LiDAR point cloud BEV maps, while the bottom row presents the reconstructed BEV maps obtained from the transmitted discrete indices. The high structural consistency indicates that the DPR module effectively preserves the geometric and semantic information of the original scenes.}
  \label{fig:opv2v_bev}
\end{figure*}

\begin{figure*}[!htbp]
  \centering
  \includegraphics[width=1.0\linewidth]{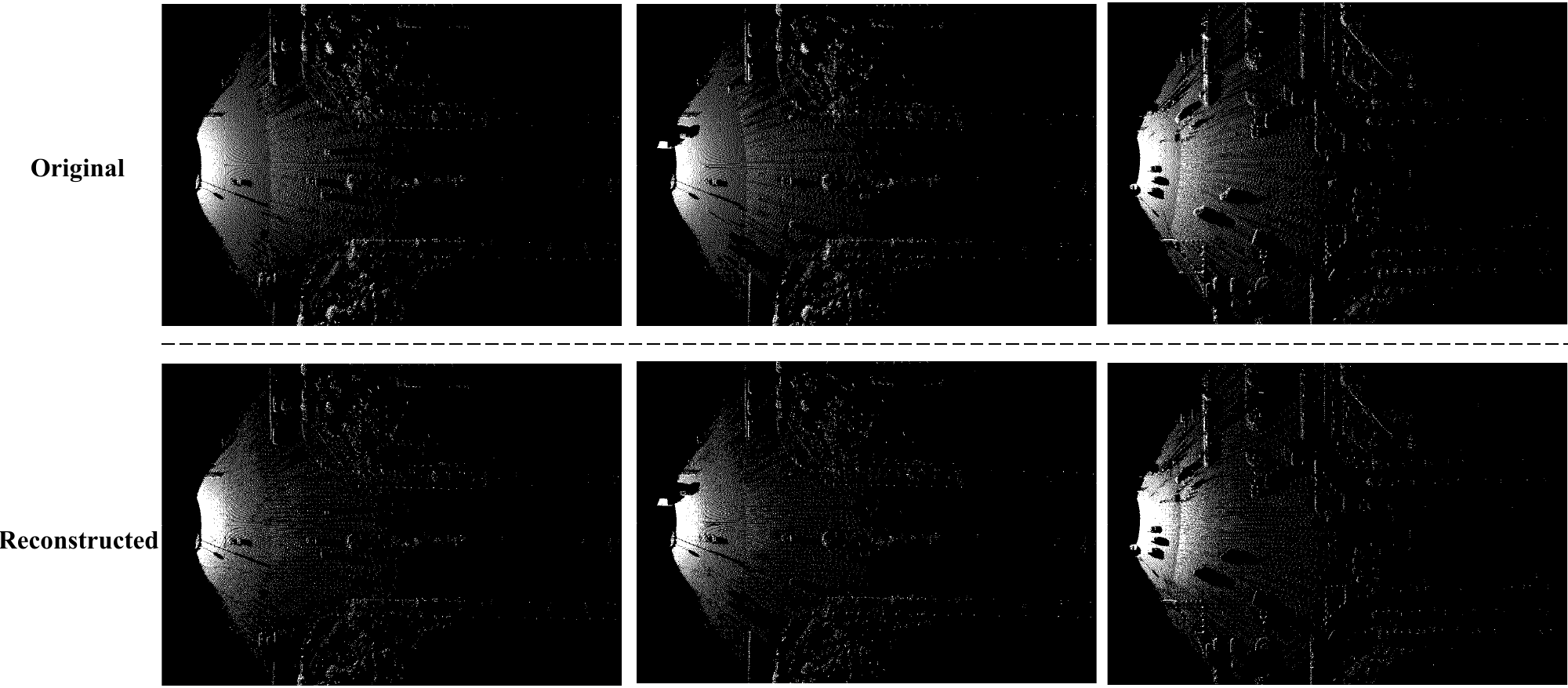}
  \caption{Visualization of BEV representations on the DAIR-V2X dataset. 
    The top row displays the original point cloud BEV maps, and the bottom row shows the reconstructed BEV maps derived from discrete indices. The reconstruction quality demonstrates the strong representation capability and information retention of the DPR module across different domains.}
  \label{fig:dairv2x_bev}
\end{figure*}

We visualize the effectiveness of the Discrete Point Cloud Representation (DPR) module by comparing top-down views of the original point clouds with those reconstructed from transmitted discrete indices. As shown in Fig.~\ref{fig:opv2v_bev} and Fig.~\ref{fig:dairv2x_bev} for the OPV2V and DAIR-V2X datasets, respectively, the reconstructed point clouds retain the spatial structure and geometric layout of the original scenes while preserving essential object information. These results highlight that DPR enables compact transmission without significant loss of critical spatial details, supporting accurate and robust collaborative perception.

\subsection{Visualization of Detection Results}
\label{sub:visual Detection}

\begin{figure*}[!htbp]
  \centering
  \includegraphics[width=1.0\linewidth]{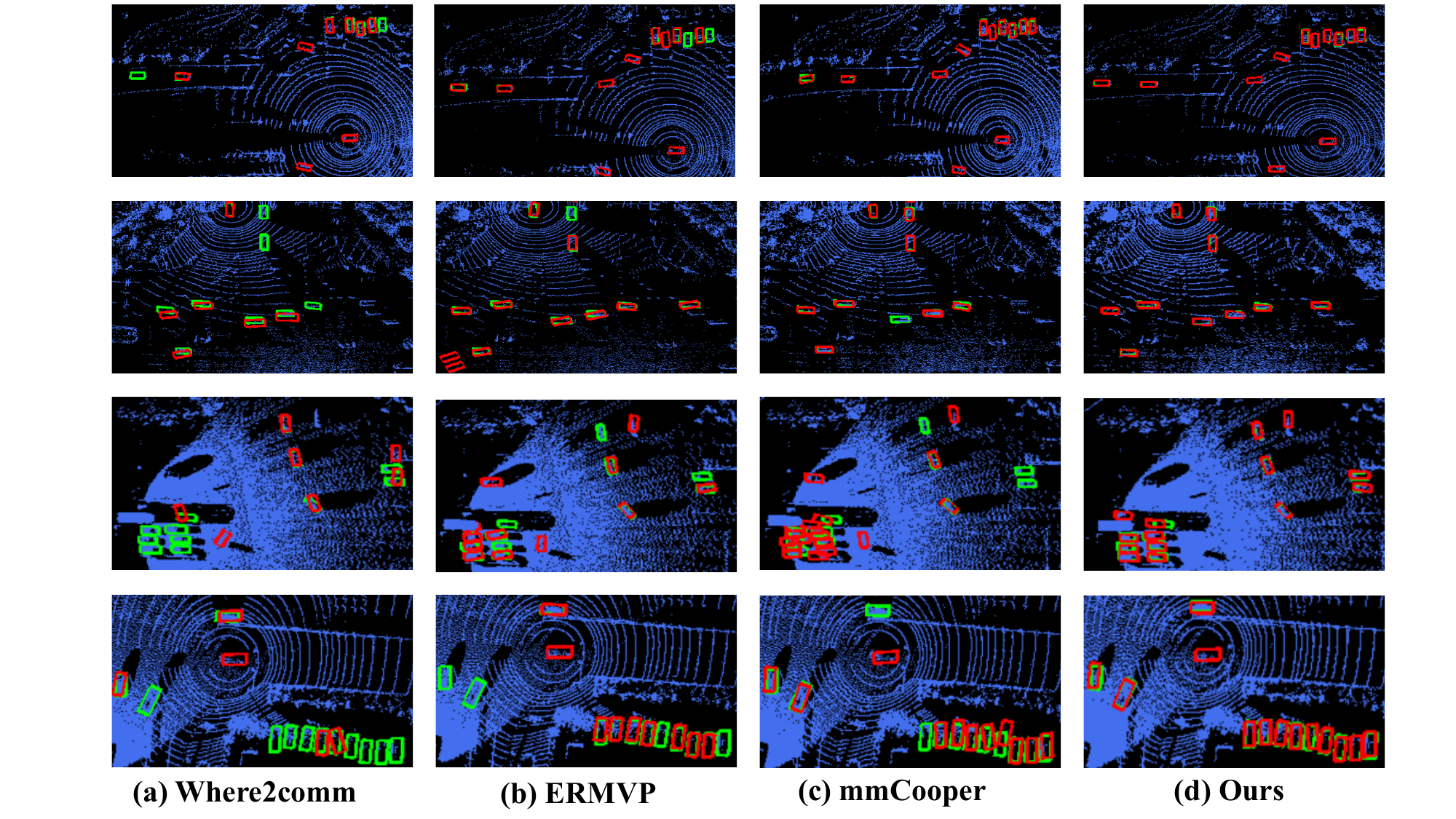}
  \caption{Additional qualitative comparison of detection results on the DAIR-V2X dataset. Green and red bounding boxes denote the ground-truth annotations and the predictions generated by the models, respectively.}
  \label{fig:visual_additional}
\end{figure*}

We further present additional qualitative results on the DAIR-V2X dataset. As shown in \cref{fig:visual_additional}, the visualizations across diverse road scenarios indicate that our proposed method achieves highly accurate object detection, successfully capturing the majority of ground-truth instances with negligible false positives. This strong performance stems from two key designs: Discrete Point Cloud Representation (DPR) preserves rich raw geometric information, enabling more faithful reconstruction and clearer object shapes after transmission, while Cascade Attention Fusion (CAF) cascades ego features before fusion, effectively compensating for occluded regions. Together, these components yield highly consistent and robust perception under complex and dynamic driving environments.

\section{Quantitative comparsion of original and reconstructed point cloud }
\label{chamfer_distance}
We report the quantitative Chamfer Distance between the original and reconstructed point clouds in \cref{tab:chamfer_loss}, indicating that the reconstruction preserves the original point cloud geometry well.
\begin{table}[!htbp]
\centering
\caption{Chamfer Distance on OPV2V and DAIR-V2X dataset.}
\label{tab:chamfer_loss}
\resizebox{0.7\textwidth}{!}{%
\footnotesize
\begin{tabular}{c|c|c}
\hline
Dataset & OPV2V & DAIR-V2X \\
\hline
Chamfer Distance (m)  & 0.0572 & 0.0516 \\
\hline
\end{tabular}
}
\end{table}

\section{Feature-level quantization vs Ours}
\label{feature_vs_ours}
Our method differs fundamentally from feature-level quantization methods (e.g., CodeFilling) in three key aspects, namely the quantization target, the reconstruction objective, and the semantic meaning of transmitted indices:
\begin{itemize}
    \item \textbf{Quantization target.} We quantize LiDAR , whereas feature-level methods quantize latent semantic features.
    \item \textbf{Reconstruction objective.} We reconstruct LiDAR with explicit spatial structure, while feature-level methods reconstruct feature maps without geometric form.
    \item \textbf{Semantic meaning of transmitted indices.} Our discrete indices are voxel-aligned and spatially grounded, carrying explicit geometric meaning, whereas feature-level methods transmit feature-semantic indices without direct geometric correspondence.
\end{itemize}
\section{More quantitave results under different settings}
\label{more_settings}
We report AP@0.7 results under varying communication latency, localization error, and heading error in \cref{tab:latency_ap07,tab:loc_error_ap07,tab:heading_error_ap07}, demonstrating that our method consistently achieves superior performance over prior approaches under all settings.

\begin{table*}[t]
\centering
\caption{AP@0.7 under different communication latency (s) on OPV2V and DAIR-V2X.}
\label{tab:latency_ap07}
\resizebox{\textwidth}{!}{%
\begin{tabular}{c|ccccc|ccccc}
\hline
 & \multicolumn{5}{c|}{\textbf{OPV2V}} & \multicolumn{5}{c}{\textbf{DAIR-V2X}} \\
\textbf{Method} 
& \textbf{0.0s} & \textbf{0.1s} & \textbf{0.2s} & \textbf{0.3s} & \textbf{0.4s}
& \textbf{0.0s} & \textbf{0.1s} & \textbf{0.2s} & \textbf{0.3s} & \textbf{0.4s} \\
\hline
Ours        & \textbf{86.81} & \textbf{81.78} & \textbf{80.66} & \textbf{80.70} & \textbf{77.92}
            & \textbf{56.50} & \textbf{53.10} & \textbf{51.67} & \textbf{51.83} & \textbf{51.63} \\
mmCooper    & 86.41 & 78.57 & 77.44 & 75.95 & 75.26
            & 56.06 & 53.05 & 50.96 & 49.86 & 50.81 \\
ERMVP       & 80.55 & 71.42 & 68.89 & 68.52 & 68.31
            & 53.27 & 49.81 & 49.68 & 48.74 & 48.60 \\
Where2comm  & 78.47 & 71.60 & 65.23 & 58.73 & 53.33
            & 52.34 & 49.74 & 47.76 & 45.83 & 45.25 \\
V2X-ViT     & 77.88 & 71.41 & 66.08 & 55.71 & 54.74
            & 46.12 & 45.69 & 45.49 & 44.54 & 43.99 \\
DiscoNet    & 77.00 & 69.20 & 63.30 & 57.40 & 53.20
            & 45.50 & 45.10 & 44.60 & 44.10 & 43.60 \\
SICP        & 71.48 & 68.23 & 67.81 & 68.01 & 67.34
            & 41.72 & 40.95 & 40.29 & 40.08 & 38.40 \\
When2com    & 70.82 & 58.87 & 43.96 & 39.90 & 36.35
            & 39.50 & 38.24 & 36.83 & 35.18 & 33.72 \\
No Fusion   & 48.66 & 48.66 & 48.66 & 48.66 & 48.66
            & 43.57 & 43.57 & 43.57 & 43.57 & 43.57 \\
\hline
\end{tabular}
}
\end{table*}

\begin{table*}[t]
\centering
\caption{AP@0.7 under different localization error (m) on OPV2V and DAIR-V2X.}
\label{tab:loc_error_ap07}
\resizebox{\textwidth}{!}{%
\small
\begin{tabular}{c|ccccc|ccccc}
\hline
 & \multicolumn{5}{c|}{\textbf{OPV2V}} & \multicolumn{5}{c}{\textbf{DAIR-V2X}} \\
\textbf{Method} 
& \textbf{0.0\,m} & \textbf{0.1\,m} & \textbf{0.2\,m} & \textbf{0.3\,m} & \textbf{0.4\,m} 
& \textbf{0.0\,m} & \textbf{0.1\,m} & \textbf{0.2\,m} & \textbf{0.3\,m} & \textbf{0.4\,m} \\
\hline
Ours        & \textbf{86.81} & \textbf{85.91} & \textbf{85.50} & \textbf{83.68} & \textbf{80.30}
            & \textbf{56.50} & \textbf{55.68} & \textbf{55.72} & \textbf{50.53} & \textbf{49.46} \\
mmCooper    & 86.41 & 84.15 & 82.35 & 79.70 & 76.80
            & 56.06 & 52.77 & 51.25 & 48.17 & 47.66 \\
ERMVP       & 80.55 & 79.07 & 76.19 & 75.46 & 72.78
            & 53.27 & 51.91 & 48.66 & 46.66 & 45.97 \\
Where2comm  & 78.47 & 77.27 & 75.45 & 72.14 & 69.77
            & 52.34 & 51.05 & 49.34 & 47.90 & 46.96 \\
V2X-ViT     & 77.88 & 76.31 & 75.21 & 70.77 & 66.72
            & 46.09 & 45.21 & 43.81 & 42.83 & 42.53 \\
DiscoNet    & 77.00 & 75.80 & 71.50 & 66.40 & 60.50
            & 45.50 & 45.10 & 44.30 & 43.20 & 41.90 \\
SICP        & 71.48 & 68.36 & 68.30 & 67.83 & 67.11
            & 41.72 & 41.35 & 40.97 & 40.58 & 39.59 \\
When2com    & 70.82 & 67.92 & 64.35 & 62.48 & 59.98
            & 39.50 & 38.06 & 36.62 & 35.84 & 35.03 \\
No Fusion   & 48.66 & 48.66 & 48.66 & 48.66 & 48.66
            & 43.57 & 43.57 & 43.57 & 43.57 & 43.57 \\
\hline
\end{tabular}
}
\end{table*}

\begin{table*}[t]
\centering
\caption{AP@0.7 under different heading errors ($^\circ$) on OPV2V and DAIR-V2X.}
\label{tab:heading_error_ap07}
\resizebox{\textwidth}{!}{%
\begin{tabular}{c|ccccc|ccccc}
\hline
 & \multicolumn{5}{c|}{\textbf{OPV2V}} & \multicolumn{5}{c}{\textbf{DAIR-V2X}} \\
\textbf{Method}
& \textbf{0.0$^\circ$} & \textbf{0.1$^\circ$} & \textbf{0.2$^\circ$} & \textbf{0.3$^\circ$} & \textbf{0.4$^\circ$}
& \textbf{0.0$^\circ$} & \textbf{0.1$^\circ$} & \textbf{0.2$^\circ$} & \textbf{0.3$^\circ$} & \textbf{0.4$^\circ$} \\
\hline
Ours        & \textbf{86.81} & \textbf{86.43} & \textbf{85.64} & \textbf{84.33} & \textbf{81.81}
            & \textbf{56.50} & \textbf{56.37} & \textbf{53.76} & \textbf{52.70} & \textbf{51.34} \\
mmCooper    & 86.41 & 84.82 & 83.18 & 81.70 & 79.91
            & 56.06 & 54.77 & 53.11 & 50.17 & 49.70 \\
ERMVP       & 80.55 & 77.07 & 74.32 & 72.46 & 71.23
            & 53.27 & 52.81 & 51.38 & 47.66 & 48.48 \\
Where2comm  & 78.47 & 77.27 & 75.45 & 72.14 & 69.77
            & 52.34 & 51.75 & 50.39 & 49.47 & 48.76 \\
V2X-ViT     & 75.88 & 75.71 & 75.69 & 74.77 & 73.89
            & 46.09 & 45.81 & 45.54 & 45.23 & 44.77 \\
DiscoNet    & 77.00 & 74.38 & 72.50 & 68.30 & 67.54
            & 45.50 & 44.45 & 43.38 & 42.25 & 41.59 \\
SICP        & 71.48 & 69.56 & 68.37 & 67.23 & 67.12
            & 41.72 & 40.43 & 41.03 & 40.16 & 39.89 \\
When2com    & 70.82 & 68.92 & 67.90 & 66.48 & 65.67
            & 39.50 & 39.06 & 38.37 & 37.32 & 36.90 \\
No Fusion   & 48.66 & 48.66 & 48.66 & 48.66 & 48.66
            & 43.57 & 43.57 & 43.57 & 43.57 & 43.57 \\
\hline
\end{tabular}
}
\end{table*}

\end{document}